\documentclass{article}

\usepackage{PRIMEarxiv}

\usepackage{authblk}
\usepackage[utf8]{inputenc} 
\usepackage[T1]{fontenc}    
\usepackage{url}            
\usepackage{booktabs}       
\usepackage{amsfonts}       
\usepackage{nicefrac}       
\usepackage{microtype}      
\usepackage{fancyhdr}       
\usepackage{graphicx}       

\usepackage[hidelinks]{hyperref}       
\usepackage{amsmath}
\usepackage[table,xcdraw]{xcolor}
\usepackage{listings}
\usepackage{array}
\usepackage{float}
\usepackage{algorithm}
\usepackage{algpseudocode}

\usepackage{multirow}
\usepackage{rotating}

\usepackage{caption}
\usepackage{subcaption}
\usepackage{verbatim}

\usepackage[toc,page]{appendix}
\usepackage{tikz}
\usetikzlibrary{shapes.geometric, arrows.meta, positioning, calc, fit}
\usepackage{pgfplots}
\usepackage{svg}
\usepackage{bbm}
\usepackage{bm}
\usepackage{natbib}

\usepackage{amssymb,mathtools}
\usepackage{xcolor}

\newcommand{\LEAP}{\textsc{Leap}}

\graphicspath{{media/}}     

\title{LEAP: a Latent Embedding Anomaly Patch-Transformer for
       time-series anomaly detection} 

\author[ ]{
 Cristhian Moya-Mota \textsuperscript{1,2},
 Ignacio Aguilera-Martos \textsuperscript{1,2},
 Diego García-Gil \textsuperscript{2,3},
 Julián Luengo \textsuperscript{1,2}
 }

\affil[1]{Department of Computer Science and Artificial Intelligence, University of Granada, Granada, Spain.}
\affil[2]{Andalusian Institute of Data Science and Computational Intelligence (DaSCI), University of Granada, Granada, Spain.}
\affil[3]{Department of Software Engineering, University of Granada, Granada, Spain}
\affil[]{\texttt{Emails: cris190402@correo.ugr.es, nacheteam@ugr.es, djgarcia@ugr.es and julianlm@decsai.ugr.es}}

\pgfplotsset{compat=1.18}
\begin{document}

\let\WriteBookmarks\relax
\def\floatpagepagefraction{1}
\def\textpagefraction{.001}

\maketitle

\begin{abstract}
Time-series anomaly detection is dominated by reconstruction-based neural detectors, whose objective is structurally flawed. Since the target is also the input, an expressive model can learn to reconstruct anomalies too, leaving uninformative residuals. Recent large-scale benchmarks show that, on curated data, elaborate deep detectors rarely beat simple statistical baselines, while current leaders rely on representation learning. We propose \LEAP{} (Latent Embedding Anomaly Patch-Transformer), which combines masked prediction with latent representation-based anomaly evidence, without relying on reconstruction or attention-based scoring. The model removes patches from a window and predicts them from the remaining temporal and cross-channel context, so prediction errors measure how well learned dynamics explain hidden segments. Deterministic complementary group masking ensures every time step is scored while hidden, with identical visible fractions during training and inference. \LEAP{} adds three label-free evidence sources: latent-embedding novelty calibrated by local density, local temporal contrast, and per-channel novelty pooled over channels selected by six label-free criteria. Under the TSB-AD protocol, the full model achieves a VUS-PR of $0.5179 \pm 0.0022$ on $180$ held-out multivariate series. On TSB-AD-U, a univariate variant combining multi-scale masked-prediction scoring with latent novelty, without contrast or channel-pool novelty, achieves $0.6194 \pm 0.0016$ on $350$ held-out series. Both results average three random seeds. A five-stage ablation on TSB-AD-M, supported by paired Wilcoxon signed-rank tests, shows statistically significant gains from each successive component.
\end{abstract}


\section{Introduction}
\label{sec:introduction}
Time-series anomaly detection (TSAD) is the task of deciding, for every instant of a recorded signal, whether the behavior observed there is compatible with the process that produced the rest of the signal \cite{blazquez2021review}. It underpins condition monitoring in industrial plants, fault diagnosis in spacecraft telemetry, intrusion detection in cyber-physical systems, arrhythmia screening in electrocardiography, and the operation of large software systems that must detect failures before their users do \cite{yan2024comprehensive}.

What distinguishes it from anomaly detection on unordered data is that the unit
of analysis is rarely a single sample. Normality depends on temporal context, which encompasses periodicity, trend, and regime. As a result, a value that is unremarkable at one phase of a cycle can become pathological shortly thereafter, indicating that the true object of abnormality is usually a subsequence: a
deformed heartbeat with the same mean, variance and range as a healthy one, a
sustained shift of a sensor's operating level, a burst of variance, or a
configuration of channels that is impossible jointly even though each channel is
plausible on its own \cite{darban2025survey}. Labels are scarce and unbalanced,
the catalog of anomaly types is heterogeneous enough that a detector tuned for
spikes fails on level shifts, and evaluation has to credit a detection that is
correct in substance but a few samples early or late.

The families of methods addressing this task have accumulated rather than replaced one another. The earliest fit a model to the series and inspect its residuals, from ARIMA and exponential smoothing onwards \cite{box2015time}. A second embeds the series into a space of sliding windows and identifies windows that are dissimilar to the rest, with the time-series discord providing the canonical formulation \cite{keogh2005hotsax}. A third learns a predictor of future values with a neural network and scores forecast residuals \cite{malhotra2015lstmad}. A fourth, dominant for much of the last decade, reconstructs its input and flags large reconstruction errors as anomalies \cite{su2019omnianomaly}, while a fifth learns short-subsequence representations through self-supervision and detects anomalies by their distance to normal representations \cite{park2026paano}.

Given its prevalence, the reconstruction family warrants closer examination. A key limitation of this approach is that the target is also used as input, a characteristic that cannot be fully addressed through architectural refinements alone. A model with
enough capacity can learn the identity map, and what it copies includes the
anomalies. The very property that makes a network expressive enough to
represent complex normal dynamics is the property that lets it reproduce the
abnormal ones. Practical systems compensate by integrating additional evidence into the score, such as inter-channel structure terms, amplitude statistics, hand-crafted shape metrics, and association discrepancies, thereby tacitly admitting that the residual alone does not carry the information the detector needs.

The practical significance of this issue is well established. Wu \cite{wu2023flawed} demonstrated that standard benchmark datasets are severely compromised by triviality, mislabeling, and run-to-failure bias.
Kim \cite{kim2022rigorous} showed that the widely used \emph{point-adjust} protocol is sufficiently permissive for random scores to outperform published detectors. Schmidl \cite{schmidl2022anomaly} and, more recently,
Liu \cite{liu2024tsbad} re-ran dozens of algorithms under controlled conditions
and reached a conclusion that should reorganize the research agenda: simple
architectures generally match or beat elaborate designs, and no single method
dominates across data sources. The numbers are blunt. On the multivariate
track of TSB-AD, which curates $1{,}070$ manually inspected series across $40$
datasets and reports VUS-PR~\cite{boniol2025vusjournal} as its primary measure, the best-performing
Transformer detector is TranAD at $0.18$ VUS-PR, while Anomaly Transformer and DCdetector, the two most cited Transformer detectors in the field, achieve scores of $0.12$ and $0.09$, respectively, falling below the $0.20$ of an Isolation Forest and far below the $0.31$ of a plain convolutional forecaster \cite{liu2024tsbad}.

Since then, the immediate context of this work has shifted. A new class of detectors has raised the multivariate state of the art from the $0.31$--$0.35$ range to approximately $0.48$ VUS-PR, largely by learning representations of short subsequences and scoring test windows by their distance to normal ones. VACE follows this principle and reaches $0.48$ VUS-PR \cite{cencillo2026vace}, while PaAno reaches $0.43$ using nearest-neighbor scoring over learned patch representations \cite{park2026paano}. KAN-AD \cite{zhou2025kanad} achieves $0.41$ through Fourier-based temporal modeling, while CrossAD \cite{li2025crossad} reaches $0.33$ using cross-scale reconstruction and cross-window context. Thus, the recent shift has been away from reconstruction-based Transformers toward representation learning with distance-based scoring. This convergence motivates our design, which combines a masked-prediction backbone with the same family of scoring rules.

Rather than mitigating the limitations of reconstruction-based detection, this paper addresses their underlying cause. We present \LEAP{} (Latent Embedding Anomaly Patch-Transformer), a semi-supervised detector whose base objective is predict-from-context rather than standard reconstruction. In this approach, the input window is divided into fixed-length patches across the temporal and channel dimensions. A deterministic subset of these patches is replaced by a learned mask embedding, requiring the network to predict the missing content using the remaining temporal and cross-channel context. As the masked patch is not visible to the encoder, the model cannot simply copy its content during reconstruction. Therefore, a large prediction error indicates that the observed context and the learned normal dynamics of the series are insufficient to explain the hidden values. Because the masking groups form an exact partition of the patch grid, every time step is predicted exactly once and
always from context, and because a single group is masked both during training
and during scoring, there is no distribution shift between how the model was
trained and how it is used.

The second contribution addresses the structural limitations of masked prediction error, which may miss sustained level shifts and globally novel but locally predictable patterns. \LEAP{} first combines masked prediction error with local moments to capture level shifts (Stage~1), then adds three complementary sources of label-free evidence: (i) density-calibrated latent-window novelty (Stages~2--3); (ii) local temporal contrast to improve precision in non-stationary recordings (Stage~4); and (iii) per-channel novelty over channels selected by six label-free criteria (Stage~5). Their additive fusion preserves complementary anomaly evidence. The five-stage ablation evaluates density calibration separately from the latent novelty it refines, rather than treating it as an additional evidence source.

The contributions of this work are the following.

\begin{enumerate}
  \item A masked-modeling detector for multivariate time-series anomaly
  detection, implemented over a Transformer encoder, with a deterministic
  complementary group masking scheme that
  staggers holes across channels, so that a broken inter-channel relationship
  surfaces in the same error signal as a deformed waveform or a level shift,
  with no hand-crafted structural metric
  (Sections~\ref{sec:leap-backbone}--\ref{sec:leap-masking}).
  \item A factorized temporal/channel encoder with an optional latent
  bottleneck on the channel axis, which reduces the cost of the cross-channel
  route from $\mathcal{O}(C^2)$ to $\mathcal{O}(CK)$ making hundreds of channels affordable
  (Section~\ref{sec:leap-backbone}).
  \item A label-free scoring pipeline (comprising per-channel error calibration with a spread floor, soft channel aggregation, edge-weighted overlap-add, event-scale smoothing, and tolerance dilation) alongside three additive evidence layers: density-calibrated latent $k$-nearest-neighbor novelty, local contrast, and criterion-based per-channel novelty (Sections~\ref{sec:leap-scoring}--\ref{sec:leap-stage5}).
  \item A five-stage decomposition of the model that doubles as a controlled
  ablation, where each stage is a strict extension of the previous one and
  reduces to it exactly by setting a single coefficient to zero
  (Section~\ref{sec:leap-stages}).
  \item An evaluation under the TSB-AD protocol on $180$ multivariate and
  $350$ univariate held-out series, with VUS-PR as the primary measure, paired
  Wilcoxon signed-rank tests for every increment, and full disclosure of the
  tuning procedure (Sections~\ref{sec:experimental} and~\ref{sec:results}).
\end{enumerate}

The remainder of the paper is organized as follows. Section~\ref{sec:background} reviews the foundations of TSAD, major methodological families, and current benchmarking evidence. Section~\ref{sec:leap} introduces the proposed \LEAP{} framework. Section~\ref{sec:experimental} details the experimental setup, and Section~\ref{sec:results} presents and discusses the results. Finally, Section~\ref{sec:conclusions} concludes this manuscript.


\section{Background}
\label{sec:background}

This section identifies the anomaly-detection mechanisms on which \LEAP{} builds and the evaluation evidence that motivates their combination.

\subsection{From Anomaly Detection to Time Series}
\label{sec:bg-ad}

Anomalies may be point-wise, contextual, or collective, depending on whether an observation is unusual in isolation, only in its context, or as part of an otherwise plausible group \cite{chandola2009}. Detection may be supervised (labeled anomalies), semi-supervised (a training sample believed to be normal), or unsupervised (a potentially contaminated sample). \LEAP{} follows the semi-supervised setting of the benchmark used here. It trains on a clean prefix and scores the complete series without using anomaly labels.

Several classical scoring rules remain relevant after features are learned. Nearest-neighbor distance measures novelty, but can over-score normal samples in sparse regions. Local-density correction, as in Local Outlier Factor, compares a sample with its neighbors on their own scale \cite{breunig2000lof}. Isolation-based methods instead use the ease of separating a sample by random splits \cite{liu2012isolation}, while subspace methods score what a low-dimensional projection fails to explain \cite{shyu2003pca}. \LEAP{} uses distance and local-density correction in its latent space.

For time series, the anomalous object is often a subsequence whose meaning depends on temporal context, rather than an isolated value. Sliding windows preserve that context and permit distance, density, or subspace scoring of temporal patterns. They also expose shape and cross-channel relationships that point-wise scoring would discard.

\subsection{Time-Series Anomaly Detection}
\label{sec:bg-tsad}

Let $X=(x_1,\dots,x_T)$, with $x_t\in R^C$, denote a series of length $T$ and $C$ channels. A detector returns a score $s_t\in R$ for each time step, with larger scores indicating stronger evidence of abnormality. In the semi-supervised setting it fits on a clean prefix $X_{1:T_{\mathrm{tr}}}$. Relevant anomalies include global or contextual points, deformed subsequences, sustained level or variance changes, altered seasonality or trend, and joint channel configurations whose individual values remain plausible \cite{blazquez2021review}. This diversity motivates complementary evidence rather than a single score.

\subsubsection{Statistical Approaches}

Classical models score departures from temporal behavior. ARIMA, exponential smoothing, and state-space models use prediction residuals. Control charts accumulate evidence of a sustained shift, and seasonal decomposition isolates departures from periodic structure \cite{box2015time}. These approaches are transparent, but a low-order or single-channel model may miss shape-preserving and relational anomalies. Applying classical methods to windows helps. Subsequence PCA \cite{shyu2003pca}, local density, and nearest-neighbor scoring are competitive on univariate benchmarks \cite{liu2024tsbad}.

\subsubsection{Discord and Bank-Based Approaches}

A time-series discord is a subsequence whose nearest non-trivial match is unusually distant \cite{keogh2005hotsax}. The Matrix Profile computes such distances across a series \cite{yeh2016matrixprofile}. Other methods compare windows with representative normal patterns. NormA builds a normal-pattern set \cite{boniol2021norma}, while shape clustering supplies normal centroids \cite{paparrizos2016kshape}. Their reusable principle is comparison with a bank of normal subsequences, although raw-space distances can be sensitive to amplitude, phase, and channel count.

\subsubsection{Deep Forecasting and Reconstruction Detectors}

Deep forecasting extends residual scoring. LSTM-AD models prediction errors \cite{malhotra2015lstmad}, DeepAnT uses a convolutional predictor \cite{munir2019deepant}, and Telemanom adds dynamic thresholding \cite{hundman2018telemanom}. Forecast error can nevertheless be high on normal but unpredictable signals. Reconstruction methods instead score the error of reproducing an observed window. Examples include OmniAnomaly's stochastic recurrent model \cite{su2019omnianomaly}, USAD's adversarial decoders \cite{audibert2020usad}, and GDN's explicit sensor dependencies \cite{deng2021gdn}. Because a reconstruction model sees its target as input, sufficient capacity may also reproduce anomalies. Its residual alone need not measure abnormality reliably. \LEAP{} addresses this risk by predicting content withheld from its encoder. Masked prediction differs from both, since it hides the target from the encoder while still using context on both sides of a gap and in other channels.

\subsubsection{Attention-Based Detectors}
\label{sec:bg-transformers}

Attention provides a way to model long temporal and cross-channel dependencies, but the scoring objective still matters. Anomaly Transformer compares learned and prior associations \cite{xu2022anomalytransformer}. TranAD combines reconstruction with self-conditioning and adversarial training \cite{tuli2022tranad}. DCdetector contrasts patch representations rather than reconstructing input \cite{yang2023dcdetector}. Their original benchmark results require caution, as common datasets contain flaws that undermine comparisons \cite{wu2023flawed}, while point-adjusted evaluation can artificially inflate detection performance \cite{kim2022rigorous}. Under the curated, range-aware TSB-AD evaluation, these published Transformer detectors rank below simpler alternatives \cite{liu2024tsbad}. This observation concerns their objectives and measured scores, not an inherent limitation of attention as an encoder.

\subsubsection{Self-Supervised Representation Learning}

Self-supervised methods learn features without labeled anomalies. TS2Vec learns multiscale contrastive representations \cite{yue2022ts2vec}. CARLA injects synthetic anomalies as contrastive negatives \cite{darban2024carla}, while COUTA uses anomaly-informed calibration of a one-class boundary \cite{xu2024couta}. Such approaches encode assumptions about future anomalies. Masked modeling instead hides part of an input and predicts it from the remainder, a principle established for image patches by masked autoencoders \cite{he2022mae}. PatchTST introduced masked patch prediction for time-series representation learning \cite{nie2023patchtst}. SimMTM reconstructs masked points by aggregating representations from multiple masked views \cite{dong2023simmtm}. MOMENT extends masked patch reconstruction to general-purpose time-series foundation models \cite{goswami2024moment}, while TSPulse employs masked reconstruction in both temporal and frequency domains \cite{ekambaram2025tspulse}. In \LEAP{}, masked prediction error serves directly as the detection score, requiring each position to be scored under masking with comparable training and inference contexts; the corresponding rules are detailed in Section~\ref{sec:leap}. Table~\ref{tab:families} in Appendix~\ref{app:background} summarizes the scoring mechanisms of these families and the anomaly types to which they are suited.

\subsubsection{Representation Learning with Nearest-Neighbor Scoring}
\label{sec:bg-recent}

Recent detectors combine a learned encoder with geometric scoring against normal embeddings. PaAno trains a compact patch encoder with a triplet and consecutiveness objective, then scores distance to a coreset bank \cite{park2026paano}. VACE regularizes embedding trajectories through velocity consistency and combines position and direction scores \cite{cencillo2026vace}. PaAno$+$ adds multiscale and cross-variable modeling \cite{zhu2026paanoplus}. PAI identifies an amplitude blind spot in window-normalized embeddings and restores raw-amplitude evidence \cite{zhang2026pai}. \LEAP{} likewise preserves level information and includes a local-moment term (Section~\ref{sec:leap-moments}).

The common design is an encoder trained on normal windows followed by a geometric anomaly score. The scoring geometry may involve a coreset distance, a trajectory-based distance, or an amplitude-aware variant. This family is closest to \LEAP{}'s latent novelty term, but the learned representation is not the whole detector here. The masked prediction error provides an independent contextual signal, while local-density calibration reduces the tendency of a raw nearest-neighbor distance to mistake legitimate sparse normal regions for anomalies. Neither improvement requires synthetic anomalies.

Other recent models refine reconstruction or forecasting instead. CrossAD predicts finer from coarser scales with prototypes \cite{li2025crossad}, CATCH models frequency-domain channel relations \cite{wu2025catch}, DADA uses adaptive bottlenecks with adversarial decoders \cite{shentu2025dada}, xLSTMAD updates the recurrent core \cite{faber2025xlstmad}, and KAN-AD replaces the deep predictor with a Kolmogorov--Arnold expansion \cite{zhou2025kanad}.

\subsection{Positioning of This Work}
\label{sec:bg-positioning}

\LEAP{} combines two scoring mechanisms. Masked modeling \cite{he2022mae} supplies an error computed only for content hidden from the encoder, while the discord and normal-bank tradition \cite{keogh2005hotsax} supplies novelty relative to training windows. Local-density calibration \cite{breunig2000lof} adjusts that novelty for unevenly populated parts of the latent space. Unlike PaAno~\cite{park2026paano} and VACE~\cite{cencillo2026vace}, whose auxiliary objectives yield geometric scores, \LEAP{} has a contextual prediction score before latent novelty is added. The Transformer implements the predictor. Table~\ref{tab:families} therefore places its scoring logic across masked modeling and representation with a normal bank. The additive formulation is detailed in Section~\ref{sec:leap-stages}. Table~\ref{tab:sota} provides the family-level comparison. Per-detector results appear in Section~\ref{sec:results-overall}.

The two signals address different questions. Masked error asks whether a local observation follows from its visible temporal and channel context. Latent novelty asks whether the shape of a larger window resembles normal training windows at all. A window can be locally predictable yet globally unfamiliar, just as a familiar overall shape can contain a brief contextual violation. Density correction then makes the novelty term relative to the structure of the normal bank instead of using one global distance scale. 
These signals are combined rather than choosing one of them for every series. Thus, attention only exchanges information among visible patches, and the anomaly evidence arises from the error on hidden patches and from a latent comparison that does not make the whole score depend on one embedding distance.

\subsection{Evaluation Paradigms and Empirical Landscape}
\label{sec:bg-benchmarks}

Reported TSAD gains depend strongly on the data and metric. Earlier benchmarks contained trivial, mislabeled, or run-to-failure cases \cite{wu2023flawed}, while point-adjusted $F_1$ can reward even random rankings \cite{kim2022rigorous}. Range-based precision and recall \cite{tatbul2018rangebased}, affiliation scores \cite{huet2022affiliation}, and VUS metrics \cite{boniol2025vusjournal} better account for anomalous events and small timing offsets. Curated re-evaluations \cite{schmidl2022anomaly} culminate here in TSB-AD, with $1{,}070$ inspected series across $40$ datasets, separated tuning and held-out sets, and VUS-PR as the primary measure \cite{liu2024tsbad}.

The metric choice matters for the claim this paper makes. Point-wise measures penalize a detector whose score rises close to a labeled interval but outside it. Point adjustment can go too far in the other direction by crediting an entire event after one hit. VUS-PR integrates precision--recall performance across temporal tolerance widths, avoiding a single chosen buffer and a single chosen decision threshold. Its precision--recall basis also reflects the rarity of anomalous time steps. The full metric definitions and secondary measures are given in Section~\ref{sec:exp-metrics}.

Table~\ref{tab:sota} summarizes published TSB-AD-M results by detector family, while Section~\ref{sec:results-overall} distinguishes reported figures from our re-measurements. The strongest representation-and-bank entry reaches $0.48$ VUS-PR. This provides empirical motivation for incorporating latent geometry into \LEAP{}, but not for treating latent geometry alone as a sufficient detector: different anomaly types can remain invisible to any single source of evidence. Since the table combines heterogeneous methods and results from different implementations and evaluation conditions, small differences should not be interpreted as definitive rankings. Section~\ref{sec:results} reports individual baselines and explicitly identifies results re-measured under the same protocol; the experimental protocol and metrics are given in Section~\ref{sec:experimental}.

\begin{table}[htbp]
\centering
\small
\caption{How the method families of Table~\ref{tab:families} stand on the
TSB-AD-M held-out evaluation set (180 series), mean VUS-PR, aggregated from the
per-detector values published in the works cited in Section~\ref{sec:bg-recent}.
\emph{Best} is the strongest published member of the family, \emph{range} the
spread across its members, and \emph{period} the publication span of those
members. The per-detector ranking is deferred to
Section~\ref{sec:results-overall}, where it serves as the comparison for
\LEAP{}.}
\label{tab:sota}
\begin{tabular}{lllcc}
\toprule
Family & Best member & VUS-PR & Range & Period \\
\midrule
Representation $+$ bank & VACE \cite{cencillo2026vace}      & 0.48 & 0.42--0.48 & 2026 \\
Forecasting (deep)      & KAN-AD \cite{zhou2025kanad}       & 0.41 & 0.31--0.41 & 2019--2025 \\
Reconstruction (deep)   & xLSTMAD-R \cite{faber2025xlstmad} & 0.37 & 0.19--0.37 & 2019--2025 \\
Discord/matrix profile  & MMPAD \cite{yeh2026mpbenchmark}   & 0.35 & ---        & 2026 \\
Statistical             & PCA \cite{shyu2003pca}            & 0.31 & 0.20--0.31 & 2003--2012 \\
Transformer             & TranAD \cite{tuli2022tranad}      & 0.18 & 0.09--0.18 & 2022--2023 \\
\bottomrule
\end{tabular}
\end{table}

\section{\LEAP{} (Latent Embedding Anomaly Patch-Transformer)}
\label{sec:leap}

\begin{figure}[!hbtp]
    \centering
    \includegraphics[width=0.85\textwidth]{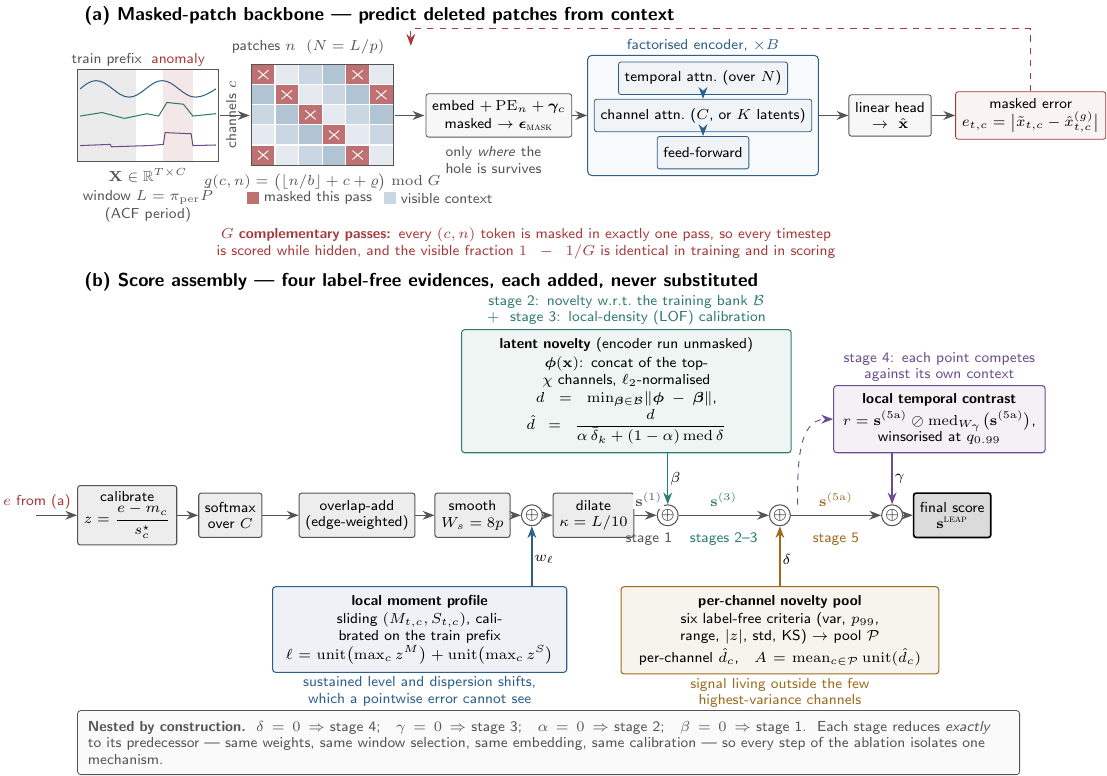}
    \caption{Architecture and scoring flow of \LEAP{}.
    Panel~(a) shows a $C\times N$ patch grid, complementary masks staggered
    across channels, the factorized encoder and masked prediction. Each patch
    is scored once while hidden. Panel~(b) shows per-channel calibration,
    overlap-add aggregation and event-scale smoothing. The score then combines
    local moments, density-calibrated latent novelty, local contrast and
    per-channel novelty from a six-criterion channel pool.}
    \label{fig:leap}
\end{figure}

This section defines the masked predictor, its score calibration and the
additional evidence used by \LEAP{}. Figure~\ref{fig:leap} shows the complete
pipeline.

\subsection{Overview and Design Principles}
\label{sec:leap-overview}

\LEAP{} combines masked prediction with representation-and-bank scoring.
Following the TSB-AD protocol \cite{liu2024tsbad}, it trains on an
anomaly-free prefix $X_{\mathrm{tr}} =
X_{1:T_{\mathrm{tr}}}\in R^{T_{\mathrm{tr}}\times C}$ and returns scores
$s\in R^T$ for the complete series $X\in R^{T\times C}$. Training,
calibration and score rescaling use no anomaly labels. Appendix~\ref{app:algorithm} gives the complete procedure in Algorithm~\ref{alg:leap}, its computational cost and a detailed account of this label-free property.

The base model masks selected patches and predicts them from visible temporal and cross-channel context, scoring each position once while hidden. Staggered gaps across channels prevent copying the input, making the resulting error a contextual prediction signal rather than a reconstruction residual. Because local prediction may miss globally unusual patterns or sustained level shifts, \LEAP{} complements it with local-moment evidence inside the base score and with three additive evidence layers: density-calibrated latent novelty, local temporal contrast, and per-channel novelty. The stages and their coefficients are defined below; notation is provided in Table~\ref{tab:notation} in Appendix~\ref{app:notation}.

\subsection{Standardization and Window Length}
\label{sec:leap-preprocessing}

The training prefix determines the channel scales and context window length for
each series.

\subsubsection{Global Per-Channel Standardization with Scale Floor}

Let $\mu_c$ and $\sigma_c$ be the mean and standard deviation of channel $c$
on $X_{\mathrm{tr}}$. Nearly constant training channels would amplify small
changes if divided by $\sigma_c$. The following scale floor uses the
median positive channel spread.
\begin{align}
  a &= \operatorname{median}\{\sigma_c : \sigma_c > 0\}, \label{eq:anchor}\\
  \varphi &= \rho_\sigma \, a, \qquad
  \mathcal{Q} = \{c : \sigma_c < \varphi\}, \label{eq:quiet}\\
  \tilde{\sigma}_c &= \max(\sigma_c, \varphi), \qquad
  \tilde{x}_{t,c} = \frac{x_{t,c}-\mu_c}{\tilde{\sigma}_c}.
  \label{eq:standardise}
\end{align}
Here $\rho_\sigma=0.05$. Using only positive spreads keeps the anchor nonzero
when most training channels are inactive. The quiescent set $\mathcal{Q}$ is
retained for calibration. Each channel uses one scale across the complete
series, which preserves level changes that per-window normalization could
remove. Reversible Instance Normalization \cite{kim2022revin} uses statistics of each window. If these include masked patches, their values leak through the normalization parameters. Per-window scaling can also erase sustained level shifts. \LEAP{} therefore uses training-prefix channel statistics throughout and predicts absolute patch levels from visible context.

\subsubsection{Dominant-Period Windowing}

The window length follows the training prefix's dominant period. For a centered
channel of length $n$, \LEAP{} computes autocorrelation $r$ by FFT, zero-padded
to $2^{\lceil\log_2 2n\rceil}$ for linear correlation and normalized to
$r(0)=1$. It selects the largest local maximum beyond lag $\ell_0=3$,
\begin{equation}
  \label{eq:period}
  \hat{P}_c = \operatorname*{arg\,max}_{\ell \in \mathcal{L}} r(\ell),
  \qquad
  \mathcal{L} = \{\ell : r(\ell) > r(\ell-1),\; r(\ell) > r(\ell+1)\},
\end{equation}
and accepted only if it clears an extreme-value band,
\begin{equation}
  \label{eq:acfband}
  r(\hat{P}_c) \;\geq\; \frac{\sqrt{2\ln m}}{\sqrt{n}},
\end{equation}
where $m=|\mathcal{L}|$. Restricting the search to peaks avoids choosing the
smallest lag of a monotonically decaying correlation. Unlike a per-lag band of $2/\sqrt{n}$, which noise peaks cross more often as more lags are searched, this band follows the Gaussian extreme-value bound for the largest of $m$ approximately $\mathcal{N}(0,1/n)$ correlations. For multivariate series, $\hat{P}$ is the median of the periods of the channels that clear the band. Estimation uses at most the $64$ highest-variance channels and the first $20{,}000$ training samples. The window is then a multiple of the patch length: 
\begin{equation}
  \label{eq:winsize}
  L = \operatorname{clip}\!\left(
      p \cdot \operatorname{round}\!\left(\frac{\pi_{\mathrm{per}} \hat{P}}{p}\right),\;
      4p,\; L_{\max}\right),
  \qquad
  L_{\max} = \min\!\left(384,\;
  p\left\lfloor \frac{T_{\mathrm{tr}}}{2p} \right\rfloor \right).
\end{equation}
With $\pi_{\mathrm{per}}=2$, the window spans about two periods. If no
significant period is found, $L=\min(L_{\max},16p)$. Multiples of $p$ need no
patch padding. The four-patch minimum leaves visible context, and the cap
bounds attention cost.

Windows use stride $\Delta=\max(1,L/8)$, reduced if needed to obtain at least
$512$ training windows. Uniform subsampling caps their number at $4{,}000$.
The tail window is always included.

\subsection{Tokenization and Encoder Architecture}
\label{sec:leap-backbone}

Patching reduces sequence length while preserving local temporal patterns \cite{nie2023patchtst}, an approach widely explored in time-series Transformers \cite{wen2023transformers}. Two-stage attention captures temporal and cross-channel dependencies efficiently \cite{zhang2023crossformer}. Given the competitiveness of linear \cite{zeng2023dlinear} and period-aware models \cite{wu2023timesnet}, we evaluate the backbone separately in Section~\ref{sec:results-sensitivity}.

\subsubsection{Tokens}

An input window $x \in R^{L \times C}$ is partitioned into a $C \times N$ grid of
patches $m{u}_{c,n} \in R^{p}$, where $n=0,\ldots,N-1$ indexes temporal position and $c$ denotes channel identity. Each patch is linearly projected into a $D$-dimensional latent space and combined with temporal and channel encodings.
The patches contain the globally standardized values from
Eq.~\eqref{eq:standardise}.
\begin{equation}
  \label{eq:tokens}
  h^{(0)}_{c,n} \;=\;
  \underbrace{\big[\,m_{c,n}\,m{\epsilon}_{\textsc{mask}}
  + (1-m_{c,n})\, (W_e m{u}_{c,n} + m{b}_e)\,\big]}_{\text{content, erased where } m_{c,n}=1}
  \;+\; \mathrm{PE}_n \;+\; m{\gamma}_c ,
\end{equation}
where $m_{c,n}\in\{0,1\}$ is the mask indicator,
$m{\epsilon}_{\textsc{mask}} \in R^{D}$ is a learned parameter,
$\mathrm{PE}_n$ is the fixed sinusoidal positional encoding of Vaswani \cite{vaswani2017attention}
and $m{\gamma}_c \in R^{D}$ is a learned channel embedding. The order of
operations in Eq.~\eqref{eq:tokens} is deliberate. The substitution happens
\emph{before} the position and channel codes are added, so what survives of a
masked token is only \emph{where} the hole is located, given by its instant and channel,
and never what was originally in it.

\subsubsection{Factorized Attention}

A multivariate window is a $C\times N$ grid of tokens. Attending over the
flattened grid would cost $\mathcal{O}((CN)^2)$, which is prohibitive on wide
datasets. \LEAP{} factorizes attention into two axes. Each of the $B$ encoder
blocks applies, in pre-norm residual form,
\begin{align}
  h &\leftarrow h + \mathrm{Attn}_{\text{time}}\big(\mathrm{LN}(h)\big),
        &&\text{over } n \text{, for each } c,
        \label{eq:tattn}\\
  h &\leftarrow h + \mathrm{Attn}_{\text{chan}}\big(\mathrm{LN}(h)\big),
        &&\text{over } c \text{, for each } n,
        \label{eq:cattn}\\
  h &\leftarrow h + \mathrm{FFN}\big(\mathrm{LN}(h)\big),
        &&\mathrm{FFN}(m{v}) = W_2\,\mathrm{GELU}(W_1 m{v}),
        \label{eq:ffn}
\end{align}
so that the cost is $\mathcal{O}(C N^2 D + N C^2 D)$ instead of
$\mathcal{O}(C^2N^2D)$.
Within one block, information can travel from $(c_1,n_1)$ to $(c_2,n_2)$
through $(c_1,n_2)$. Temporal attention supplies within-channel context.
Channel attention supplies simultaneous observations from other variables,
allowing relational disruptions to affect masked prediction. Fused
scaled dot-product kernels avoid materializing attention matrices in GPU
memory.

\subsubsection{Latent Channel Bottleneck}
When $C>C_{\text{lat}}=128$, \LEAP{} replaces quadratic channel attention
with $K$ learned channel latents
$\mathbf{\Lambda}\in\mathbb{R}^{K\times D}$, following the inducing-point
idea of Lee \cite{lee2019settransformer}. They aggregate channel information
and broadcast it back through two cross-attention steps:
\begin{align}
  \mathbf{\Lambda}' &= \mathbf{\Lambda} + \mathrm{CrossAttn}\big(Q=\mathbf{\Lambda},\; KV=\mathrm{LN}(\mathbf{h}_{:,n})\big),
  \label{eq:write}\\
  \mathrm{Attn}_{\text{chan}}(\mathbf{h}_{:,n}) &= \mathrm{CrossAttn}\big(Q=\mathrm{LN}(\mathbf{h}_{:,n}),\; KV=\mathrm{LN}(\mathbf{\Lambda}')\big),
  \label{eq:read}
\end{align}
where $\mathrm{LN}(\cdot)$ denotes layer normalization applied prior to attention. Equation~\eqref{eq:write} condenses $C$ channel representations into $K$ latent modes, while Eq.~\eqref{eq:read} distributes this aggregated cross-channel context back to individual channels. 

The channel-attention cost per patch falls from $\mathcal{O}(C^2D)$ to
$\mathcal{O}(CKD)$. With $C=248$ and $K=16$, this is roughly an eight-fold
reduction in channel-attention operations. Gradient checkpointing can further
reduce training memory.

After the final encoder block, a linear head
$W_d\in\mathbb{R}^{D\times p}$ predicts each patch from
$\mathbf{h}_{c,n}^{(B)}$. The patches form
$\hat{x}\in\mathbb{R}^{L\times C}$.

\subsection{Complementary Group Masking}
\label{sec:leap-masking}

The mask is a deterministic partition of the $(c,n)$ token grid into $G$
groups
\begin{equation}
  \label{eq:groups}
  g(c,n) \;=\; \Big( \big\lfloor n/b \big\rfloor + c + \varrho \Big) \bmod G ,
  \qquad
  Mask^{(g)} = \{(c,n) : g(c,n) = g\}.
\end{equation}
Here $b$ is the block length in patches, giving gaps of $bp$ samples, and
$\varrho$ is a rotation offset. Evaluating all $G$ groups makes the assembled
score independent of that offset. The partition has three properties.

\begin{enumerate}
  \item \textbf{Exact partition.} Every token belongs to exactly one group, so
  masking the $G$ groups in turn predicts every token exactly once and always
  from context. There is no position in the output that the model was allowed
  to see while predicting it.
  \item \textbf{Train--inference consistency.} Training and inference each
  mask one group per pass. Pass $g$ leaves a fraction
  $1-|\mathcal{M}^{(g)}|/(CN)$ visible, averaging $1-1/G$ over all groups in
  both phases. Thus a masked prediction is evaluated with the same
  single-group context pattern used for training.
  \item \textbf{Staggering across channels.} The additive $c$ term rotates the
  pattern by one group per channel. Without it, a group would consist of whole
  time columns, removing every channel at the same instant and leaving the
  cross-channel route with nothing to read. The task would collapse to pure
  temporal interpolation. With it, any masked token retains both a temporal
  context and a set of simultaneous observations in other channels.
\end{enumerate}

At inference, the masked error of a window is assembled from the $G$
complementary passes,
\begin{equation}
  \label{eq:maskederror}
  e_{t,c} \;=\; \big| \tilde{x}_{t,c}
  - \hat{x}^{\,(g(c,\lfloor (t-1)/p \rfloor))}_{t,c} \big|,
\end{equation}
where $\hat{x}^{(g)} = f_\theta(\tilde{x}, Mask^{(g)})$ is the prediction
of the pass that hid the group containing $(c,\lfloor (t-1)/p\rfloor)$.

\subsection{Training Objective}
\label{sec:leap-training}

Training minimizes the mean absolute error restricted to the masked positions
of one randomly drawn group,
\begin{equation}
  \label{eq:loss}
  \mathcal{L}(\theta) \;=\;
  E_{x \sim \mathcal{D}_{\mathrm{tr}},\, g,\varrho \sim \mathcal{U}\{0,G-1\}}
  \left[
  \frac{\sum_{(c,n)\in Mask^{(g)}} \sum_{j=1}^{p}
  \big| \tilde{x}_{np+j,c} - \hat{x}_{np+j,c} \big|}
  {p\,|Mask^{(g)}|}
  \right].
\end{equation}
The $\ell_1$ loss limits the influence of contamination in a training prefix
that is assumed, but not guaranteed, to be anomaly-free. A squared loss
would let a single large residual dominate the gradient.

Optimization uses AdamW, gradient clipping, a \texttt{ReduceLROnPlateau} scheduler, and early stopping on the last $20\%$ of the training prefix. A fixed validation-mask seed keeps validation loss independent of mask-sampling noise. BF16 mixed precision is optionally supported.

\subsection{Calibration and Base Anomaly Scoring}
\label{sec:leap-scoring}

The masked error of Eq.~\eqref{eq:maskederror} is calibrated per channel,
aggregated across channels and overlapping windows, and smoothed before it
enters the Stage-1 score. Local moment evidence complements this error.

\subsubsection{Per-Channel Error Calibration}

Channels differ in how predictable they are, and a raw error is therefore not
comparable across them. \LEAP{} estimates the first two moments of the masked
error on the (anomaly-free) training windows,
\begin{equation}
  \label{eq:calibmoments}
  m_c = E[e_{\cdot,c}], \qquad
  s_c = \sqrt{E[e_{\cdot,c}^2] - m_c^2},
\end{equation}
and regularizes the spread through optional pooling toward the median of
active channels and a lower floor,
\begin{align}
  \bar{s} &= \operatorname{median}_{c\notin\mathcal{Q}} s_c, \qquad
  \tilde{s}_c = (1-\lambda)\,s_c + \lambda\,\bar{s},
  \label{eq:shrink}\\
  s^{\star}_c &= \max\!\big(\tilde{s}_c,\; \max(\rho_f\,\bar{s},\, \varepsilon)\big),
  \qquad
  z_{t,c} = \frac{e_{t,c} - m_c}{s^{\star}_c}.
  \label{eq:zscore}
\end{align}
The floor prevents an unbounded $z$ when $s_c\approx0$. The parameter
$\lambda$ ranges from the raw spread at $0$ to a pooled spread at $1$.
The reported setting $\lambda=0$ keeps the floor while disabling pooling.

\subsubsection{Channel Aggregation}

The calibrated errors are reduced over channels by a temperature-controlled
soft maximum:
\begin{equation}
  \label{eq:chanagg}
  a_{t} \;=\; \sum_{c} w_{t,c}\,\max(z_{t,c},0),
  \qquad
  w_{t,c} = \frac{\exp(z_{t,c}/\tau)}{\sum_{c'}\exp(z_{t,c'}/\tau)}.
\end{equation}
With $\tau=0.5$, the soft maximum balances the mean, which can dilute a
single-channel anomaly, against the maximum, which can favor a noisy
channel. Hard \texttt{max} and \texttt{mean} remain available.

As $\tau$ increases, weights approach uniformity; as $\tau \to 0$, the largest calibrated error dominates. The positive part in Eq.~\eqref{eq:chanagg} prevents small errors from canceling evidence across channels.

\subsubsection{Overlap-Add, Event-Scale Smoothing, and Tolerance Dilation}

Windows overlap, so a time step receives one score per window that covers it.
Positions near a window edge have seen less context and are down-weighted
by a Gaussian profile before accumulation.
\begin{equation}
  \label{eq:overlapadd}
  s_t \;=\;
  \frac{\sum_{(w,j)\,:\,\text{start}_w + j = t} \omega_j\, a_{w,j}}
       {\sum_{(w,j)\,:\,\text{start}_w + j = t} \omega_j},
  \qquad
  \omega_j = \exp\!\left(-\tfrac{1}{2}
  \left(\frac{j - L/2}{\rho_\omega L}\right)^{\!2}\right),
\end{equation}
with $\rho_\omega = 0.25$. The result is then smoothed with a moving average of
width $W_s = \eta\,p$,
\begin{equation}
  \label{eq:smooth}
  s \leftarrow smooth(s; W_s), \qquad W_s = \eta \, p, \quad \eta = 8.
\end{equation}
This reads error at event scale and yields the largest base-score gain in the
component analysis (Appendix~\ref{sec:appendix-ablation-micro}). A maximum filter of
width $\kappa=\max(3,\lfloor L/10\rfloor)$ then dilates the combined
Stage-1 signal,
\begin{equation}
  \label{eq:dilation}
  (dil_\kappa\bm{v})_t = \max_{|u| \leq \kappa/2} v_{t+u},
\end{equation}
which tolerates small temporal offsets (Section~\ref{sec:exp-metrics}).

\subsubsection{Local Moment Evidence}
\label{sec:leap-moments}

To detect sustained level or dispersion changes, \LEAP{} computes local
moments on the signal rather than inside a normalization layer. Let $M_{t,c}$
and $S_{t,c}$ be the mean and standard deviation in a centered window of
$W_m=\max(8,4p)$ samples. Cumulative sums compute them in
$\mathcal{O}(T)$, with edge replication to avoid artificial endpoint shifts.
Training-prefix baselines $(\mu^M_c,\sigma^M_c)$ and
$(\mu^S_c,\sigma^S_c)$ use a positive-median scale floor. The evidence
combines two-sided deviations in both moments,
\begin{equation}
  \label{eq:level}
  \ell_t \;=\;
  unit\!\Big(\max_c \big| M_{t,c}-\mu^M_c \big| / \sigma^M_c \Big)
  \;+\;
  unit\!\Big(\max_c \big| S_{t,c}-\mu^S_c \big| / \sigma^S_c \Big),
  \qquad
  unit(\bm{v}) = \frac{\bm{v}}{q_{0.99}(\bm{v}\mid \bm{v}>0)}.
\end{equation}

Taking absolute deviations captures both increases and decreases, and summing the two profiles keeps level and dispersion evidence instead of only the larger of them. The operator $unit$ divides each profile by the 99th percentile of its active values ($\bm{v}>0$), so that a mostly silent signal does not shrink the scale, and it adds no $\epsilon$ to the denominator, since an additive floor could hide genuine anomalies. A profile without active values is left unchanged.

The per-series rescaling changes neither profile's internal order, although their sum can produce a new ranking, and it leaves the two profiles on comparable scales. The maximum over channels lets a shift confined to a single sensor carry the evidence.

\subsubsection{Stage-1 Base Anomaly Score}
\label{sec:leap-stage1}

Stage~1 adds rescaled local moments to the smoothed prediction error before
dilation,
\begin{equation}
  \label{eq:stage1}
  s^{(1)} \;=\; dil_\kappa\Big(
  unit\big(s\big) \;+\; w_\ell \,unit(\bm{\ell})
  \Big),
  \qquad w_\ell = 1 .
\end{equation}
At $w_\ell=0$ the score uses prediction error alone. Alternative moment
combinations are compared in Section~\ref{sec:results-sensitivity}.

\subsection{Additive Evidence Layers}
\label{sec:leap-stages}

Stage~1 is complemented by latent novelty, density calibration, local contrast, and per-channel novelty. These components form five nested ablation stages, with density calibration evaluated separately from latent novelty (Appendix~\ref{sec:appendix-ablation-stages}).

\subsubsection{Latent Shape Novelty (Stage~2)}
\label{sec:leap-stage2}

A window can have an unfamiliar shape even when its patches are locally
predictable. Stage~2 measures its novelty in the encoder's latent space. Let
\begin{equation}
  \label{eq:encode}
  h(x) \;=\; \mathrm{Enc}\big(W_e\,\mathrm{patchify}(x) + \bm{b}_e
  + \mathrm{PE} + \bm{\gamma}\big) \in R^{C\times N\times D}
\end{equation}
denote the encoder representation without masking. Let $\mathcal{A}$ contain
the $\min(C,\chi)$ highest-variance training channels, with $\chi=4$.
Concatenating their patch representations and applying $\ell_2$
normalization gives
\begin{equation}
  \label{eq:embed}
  \bm{\phi}(x) \;=\;
  \frac{\operatorname{vec}\big(h_{\mathcal{A},:,:}(x)\big)}
        {\big\|\operatorname{vec}\big(h_{\mathcal{A},:,:}(x)\big)\big\|_2 + \epsilon}
  \;\in\; R^{|\mathcal{A}| N D}.
\end{equation}
Concatenation preserves temporal order that pooling would discard. Two
windows can have similar average features yet differ in the order of
their patch-level changes. Retaining that order makes shape deformation
visible to the bank distance. Because this vector is $\ell_2$-normalized, the bank distance depends on the direction of the representation, while amplitude information is retained by the separate local-moment term of Section~\ref{sec:leap-moments}.

A uniformly sampled bank $\mathcal{B}$ holds at most
$N_{\mathrm{bank}}=1200$ training-window embeddings. A test window's
novelty is its distance to the nearest bank element,
\begin{equation}
  \label{eq:morph}
  d(x) \;=\; \min_{\bm{\beta} \in \mathcal{B}}
  \big\| \bm{\phi}(x) - \bm{\beta} \big\|_2 .
\end{equation}
This is analogous to an \emph{AB-join} against a normal reference series
\cite{yeh2016matrixprofile}. Gaussian overlap-add from
Eq.~\eqref{eq:overlapadd} maps window distances to $\tilde d_t$.
Stage~2 adds them to the base score,
\begin{equation}
  \label{eq:stage2}
  s^{(2)} \;=\; unit\big(s^{(1)}\big)
  \;+\; \beta\, unit\big(\tilde{\bm{d}}\big),
  \qquad \beta = 1 .
\end{equation}

\subsubsection{Density-Based Novelty Calibration (Stage~3)}
\label{sec:leap-stage3}

Raw distance can be high in a valid but sparsely sampled normal mode.
Following the local-density idea of LOF \cite{breunig2000lof},
Stage~3 scales novelty by the bank's nearby spacing. Let
$\delta_{\bm{\beta}}$ be the leave-one-out nearest-neighbor distance
of bank element $\bm{\beta}$, and let $\mathcal{N}_k(x)$ contain the
$k$ nearest bank elements. Calibrated novelty is defined in Eq.~\eqref{eq:lof}, with $k=4$ and $\alpha=0.25$.
\begin{equation}
  \label{eq:lof}
  \hat{d}(x) \;=\;
  \frac{d(x)}
       {\alpha \cdot \overline{\delta}_k(x)
        + (1-\alpha)\cdot \operatorname{median}_{\bm{\beta}\in\mathcal{B}} \delta_{\bm{\beta}}},
  \qquad
  \overline{\delta}_k(x) = \frac{1}{k}\!\!\sum_{\bm{\beta}\in\mathcal{N}_k(x)}\!\! \delta_{\bm{\beta}}.
\end{equation}

The global median prevents the denominator from collapsing when local values approach zero. At $\alpha=0$, the calculation is proportional to raw novelty, preserving its ranking after rescaling. Although $\alpha\in[0,1]$ is recommended to maintain a positive denominator, larger values are supported through a small positive numerical floor that prevents zero or negative denominators. The deployed value $\alpha=0.25$ provides a conservative calibration. The Stage-3 score is:
\begin{equation}
  \label{eq:stage3}
  s^{(3)} \;=\; unit\big(s^{(1)}\big)
  \;+\; \beta\, unit\big(\hat{\tilde{\bm{d}}}\big).
\end{equation}

Local spacing is large in a sparse but valid normal region and small in
a densely sampled one. Dividing by that spacing prevents a legitimate
sparse mode from appearing novel solely because its neighbors are far
apart. The global component limits unstable ratios when several bank
embeddings nearly coincide.

Simple unit rescaling changes the scale of raw distance but does not
remove a high-distance plateau caused by sparse sampling. Density
calibration changes the relative values of windows in different parts
of the normal manifold. It therefore addresses a distinct problem from
the per-series scaling used for score fusion.

\subsubsection{Local Temporal Contrast (Stage~4)}
\label{sec:leap-stage4}
On long non-stationary series, a global score baseline can drift.
Stage~4 compares each score with a running median over $W_\gamma$
samples and adds the resulting contrast, defined in Eq.~\eqref{eq:contrast}.
\begin{equation}
  \label{eq:contrast}
  r_t \;=\; \frac{s^{(3)}_t}
  {med_{W_\gamma}\!\big(s^{(3)}\big)_t
   + \rho_\gamma \operatorname{median}(s^{(3)}) + \epsilon},
  \qquad
  s^{(4)} \;=\; s^{(3)} + \gamma \cdot unit_{q}\!\big(\bm{r}\big).
\end{equation}
Here $med_W$ is a running median, $unit_q$ bounds the tail at the same
quantile of the active support that $unit$ divides by --- the bound and the
divisor are one quantity, and the clip is applied only once that scale is
known to be usable --- and $\rho_\gamma=10^{-3}$ stabilizes the denominator.
The stated coefficient is $\gamma=1.5$.
At $\gamma=0$, Stage~4 equals Stage~3.

The ratio $r_t$ highlights a point that rises above its own neighborhood on a non-stationary recording. Adding it, instead of multiplying the score by it, keeps the Stage-3 signal when a long anomaly also raises its local median, which a multiplicative correction would erase. The contrast is computed on the complete Stage-3 score, so it responds to masked error and latent novelty together. Winsorization limits the effect of near-zero denominators, and $W_\gamma=\nu L$ ties the context width to the window length.

\subsubsection{Criterion-Based Channel Selection and Novelty Scoring (Stage~5)}
\label{sec:leap-stage5}

The high-variance channels of Eq.~\eqref{eq:embed} can miss anomalies in
quiet channels. Stage~5 therefore takes the union $\mathcal{P}$ of the
top $\min(C,\chi')$ channels selected by each of six label-free criteria. Let
$\bm{a} = X_{1:T_{\mathrm{tr}}}$ be the training part and
$\bm{b} = X_{T_{\mathrm{tr}}+1:T}$ the remainder. The criteria are
\begin{equation}
  \label{eq:criteria}
  \begin{aligned}
    \textsc{var}_c &= \operatorname{std}(\bm{a}_c),
    &\textsc{p99}_c &= \tfrac{1}{T}\textstyle\sum_t \mathbb{1}\!\left[x_{t,c} > q_{0.99}(\bm{a}_c)\right],\\
    \textsc{range}_c &= \frac{\operatorname{ptp}(\bm{b}_c)+\epsilon}{\operatorname{ptp}(\bm{a}_c)+\epsilon},
    &\textsc{zmean}_c &= \tfrac{1}{T}\textstyle\sum_t \frac{|x_{t,c}-\overline{a}_c|}{\operatorname{std}(\bm{a}_c)+\epsilon},\\
    \textsc{sstd}_c &= \frac{\operatorname{std}(\bm{b}_c)+\epsilon}{\operatorname{std}(\bm{a}_c)+\epsilon},
    &\textsc{ks}_c &= \sup_{v} \big| F_{\bm{a}_c}(v) - F_{\bm{b}_c}(v) \big| .
  \end{aligned}
\end{equation}
The last criterion is a two-sample Kolmogorov--Smirnov statistic
\cite{massey1951ks} computed on a fixed-seed subsample. The criteria use
the known split point $T_{\mathrm{tr}}$ and no anomaly labels.

The variance criterion reflects training activity. The remaining
criteria capture high-tail frequency, range, mean absolute deviation,
dispersion and distribution changes between the training prefix and
the unlabeled series. Because several inspect the complete remainder
$\bm b$, channel selection is an offline step. Their union allows a
low-variance channel to enter the pool even when it was absent from the
concatenated embedding of Stage~2.

For each $c\in\mathcal{P}$, the normalized embedding
$\bm{\phi}_c(x)=\operatorname{vec}(h_{c,:,:})/
\big(\|\operatorname{vec}(h_{c,:,:})\|_2+\epsilon\big)
\in R^{ND}$ is scored by the density-calibrated novelty of
Eq.~\eqref{eq:lof}, using a separate bank $\mathcal{B}_c$ of that channel's
training-window embeddings and their leave-one-out distances. Overlap-add projects it to time steps. The rescaled channel scores are averaged,
\begin{equation}
  \label{eq:agg}
  A_t \;=\; \frac{1}{|\mathcal{P}|} \sum_{c \in \mathcal{P}}
  unit\big(\hat{\tilde{d}}_{c}\big)_t .
\end{equation}
The pool evidence enters before contrast is recomputed on the combined
score:
\begin{align}
  s^{(5\text{a})} &= s^{(3)} + \delta \cdot unit(\bm{A}), \qquad \delta = 1.5,
  \label{eq:stage5a}\\
  s^{\LEAP} &= s^{(5\text{a})}
  + \gamma \cdot unit_q\!\Big(s^{(5\text{a})} \oslash
  \big(med_{W_\gamma}(s^{(5\text{a})}) + \rho_\gamma \operatorname{median}(s^{(5\text{a})}) + \epsilon\big)\Big).
  \label{eq:stage5}
\end{align}

Here $\oslash$ denotes element-wise division. The pool supplements the concatenated embedding, as replacing it degraded individual-series performance. At $\delta=0$, Eq.~\eqref{eq:stage5} reduces to $s^{(4)}$. Scoring each pooled channel independently preserves its novelty when only one sensor departs from normal patterns, helping series where the high-variance channels in Eq.~\eqref{eq:embed} obscure the anomaly. Contrast is recomputed after adding this term to capture the complete Stage-5 evidence.

\subsubsection{Five-Stage Score Summary}

Table~\ref{tab:stages} summarizes the five scores and the coefficient that
removes each successive mechanism for ablation. Further configuration details
are provided in Section~\ref{sec:exp-hyperparameters}.

\begin{table}[htbp]
\centering
\small
\caption{The five \LEAP{} scores and the coefficient that removes each
successive mechanism for ablation. Setting $\beta=0$ preserves the
Stage-1 ranking after unit rescaling.}
\label{tab:stages}
\begin{tabular}{clll}
\toprule
Stage & Integrated Mechanism & Score & Reduction Condition \\
\midrule
1 & Masked-patch prediction $+$ local moments & Eq.~\eqref{eq:stage1} & --- \\
2 & Latent shape novelty & Eq.~\eqref{eq:stage2} & $\beta = 0$ \\
3 & Local-density calibration of novelty & Eq.~\eqref{eq:stage3} & $\alpha = 0$ \\
4 & Local temporal contrast & Eq.~\eqref{eq:contrast} & $\gamma = 0$ \\
5 & Per-channel novelty over criterion pool & Eq.~\eqref{eq:stage5} & $\delta = 0$ \\
\bottomrule
\end{tabular}
\end{table}


\section{Experimental Framework}
\label{sec:experimental}

This section describes the benchmarks, the model configuration, the evaluation metrics, the tuning protocol and the computational environment, so that every reported result can be reproduced.

\subsection{Benchmarks}
\label{sec:exp-benchmarks}

Experimental evaluation uses the TSB-AD benchmark \cite{liu2024tsbad}, comprising $1{,}070$ curated time series from $40$ public datasets with human-verified anomaly labels. It includes $870$ univariate series (TSB-AD-U) and $200$ multivariate series (TSB-AD-M), with separate tuning and held-out evaluation subsets. We use $48/350$ tuning/evaluation series from $23$ datasets and nine application domains for TSB-AD-U, and $20/180$ from $17$ datasets and six domains for TSB-AD-M. The detailed composition of the evaluation sets is reported in Table~\ref{tab:benchmarks} in Appendix~\ref{app:composition-per-dataset}.

The evaluation sets are deliberately heterogeneous: TSB-AD-M spans $2$--$248$ channels and from a few hundred to over half a million time steps, while TSB-AD-U covers a broad range of univariate signals, including web-service monitoring traces. Each series specifies an anomaly-free training prefix. Semi-supervised detectors, including \LEAP{}, are trained exclusively on this prefix and then score the complete series, whereas unsupervised methods operate directly on the full series. Ground-truth labels are never used for fitting or scoring, and metrics are computed per series and averaged uniformly so that each series contributes equally regardless of length.

\subsection{Hyperparameters}
\label{sec:exp-hyperparameters}
Table~\ref{tab:hp} in Appendix~\ref{app:hyperparameters} lists the complete configuration. Three groups can be distinguished. \emph{Structural} hyperparameters ($p$, $D$, $H$, $B$, $G$) define the architecture and are fixed a priori at conventional values rather than searched. \emph{Adaptive} hyperparameters ($L$, $\Delta$, $W_s$, $W_m$, $\kappa$, $W_\gamma$) are instead derived per series from the estimated dominant period or patch length, allowing a single configuration to accommodate series with sampling rates that differ by orders of magnitude.
\emph{Fusion} hyperparameters
($w_\ell,\beta,\alpha,k,\gamma,\delta,\chi,\chi'$) are the coefficients of the
additive evidence layers and are the only ones selected on the tuning subset
(Section~\ref{sec:exp-tuning}). The architecture uses patches of $p=16$ samples, width $D=128$, $H=8$ attention heads, $B=4$ encoder blocks and $G=4$ masking groups with blocks of $b=1$ patch, and the fusion coefficients take the values given with their equations in Section~\ref{sec:leap}.

A unified hyperparameter configuration is used across all $180$ multivariate and, separately, all $350$ univariate evaluation series, without per-series or per-dataset tuning. The univariate configuration (Table~\ref{tab:hp}) adapts \LEAP{} to single-channel inputs ($C=1$) in three ways.

First, the event-scale smoothing of Eq.~\eqref{eq:smooth} is replaced by a multi-scale readout: masked-prediction errors are smoothed at widths $W_s \in \{1,8,32,128\}$ and local moments (Section~\ref{sec:leap-moments}) are computed over $W_m \in \{16,64,256\}$. Each scale is rescaled to unit range, the maximum response is retained per time step, and tolerance dilation is applied. This avoids a fixed width, whose optimum varied independently of series length, window size, and training duration.

Second, local contrast is disabled ($\gamma=0$), as it adds no signal beyond multi-scale scoring. Third, channel-pool novelty is disabled ($\delta=0$), since the pool in Eq.~\eqref{eq:criteria} reduces to one channel, making Stage~5 redundant with Stage~3. Latent novelty thus embeds that channel rather than the cross-channel concatenation of Eq.~\eqref{eq:embed}, with density calibration unchanged.

All experiments use three fixed random seeds ($2024$, $2025$, and $2026$), consistently applied to PyTorch, NumPy, Python's \texttt{random} module, and the masking generators. Deterministic cuDNN is enabled where supported, with computations performed in FP32 by default.

\subsection{Performance Metrics}
\label{sec:exp-metrics}

The evaluation adopts the TSB-AD metric suite without modification and reports nine metrics per series.

VUS-ROC and VUS-PR \cite{boniol2025vusjournal} are threshold-free and range-aware. A buffer of width $w$ is placed around each labeled anomalous region, inside which a detection receives a weight that decays with its distance to the region. The ROC or PR curve is computed for every $w \in \{0,\dots,2\ell_{\mathrm{vus}}\}$ and the resulting surface is integrated,
\begin{equation}
  \label{eq:vus}
  \mathrm{VUS\text{-}PR} \;=\;
  \frac{1}{2\ell_{\mathrm{vus}}}\int_{0}^{2\ell_{\mathrm{vus}}}\!\! \mathrm{AUC\text{-}PR}(w)\,\mathrm{d}w ,
\end{equation}
so the measure depends on neither the threshold nor the buffer width. The half-width $\ell_{\mathrm{vus}}$ is the dominant period of the series, estimated from the autocorrelation of the first channel with the peak-based rule of the benchmark and capped at $125$ when the estimate exceeds $300$. Following Liu \cite{liu2024tsbad}, VUS-PR is the primary measure of this paper. It is threshold-free, tolerant to small temporal displacements and sensitive to class imbalance, and it showed the strongest correlation with human judgment in the meta-evaluation of the benchmark.

The remaining seven metrics are reported for comparability with the benchmark. AUC-ROC and AUC-PR are point-wise and threshold-free, so a detection shifted by a few samples is fully penalized, and AUC-PR is the more informative of the two because anomalies are rare. Standard-F1 is the point-wise $F_1$ at the oracle threshold, and PA-F1 counts a whole labeled segment as detected when any of its points is flagged. Event-based-F1 treats each anomalous segment as one unit, R-based-F1 \cite{tatbul2018rangebased} rewards both the existence and the overlap of detections, and Affiliation-F \cite{huet2022affiliation} relies on the temporal distance between predicted and labeled events. Following Kim \cite{kim2022rigorous}, PA-F1 is treated as diagnostic only, since even a random score can beat a trained detector under it, and no claim in this paper rests on it.

\subsection{Tuning Strategy}
\label{sec:exp-tuning}

The tuning protocol was designed to make the reported held-out figures
defensible under the criticisms reviewed in Section~\ref{sec:bg-benchmarks}.
Five rules were enforced.

\paragraph{Strict Separation of Tuning and Evaluation Sets}
All architectural choices and hyperparameter selections were conducted exclusively on the benchmark's designated tuning subsets ($20$ multivariate and $48$ univariate series) or internal validation splits, without accessing held-out evaluation sequences. The held-out evaluation sets were reserved strictly for generating the final benchmark results in Section~\ref{sec:results}.

\paragraph{Stage-Wise Greedy Selection}
Each stage in the ablation hierarchy (Table~\ref{tab:stages}) was fixed sequentially before introducing subsequent components, searching only the scaling coefficient governing the newly integrated mechanism. This greedy optimization strategy constrains search-space complexity to scale linearly with the number of stages rather than exponentially. While higher-order parameter interactions are not explicitly explored, this trade-off is justified because each stage incorporates a physically motivated domain mechanism rather than an unconstrained hyperparameter.

\paragraph{Paired Hypothesis Testing}
Because per-series VUS-PR distributions are paired and non-Gaussian, architectural modifications were evaluated using two-sided Wilcoxon signed-rank tests computed over per-series differences. Wins, losses, and ties were tracked alongside mean differences. A modification was accepted only if it yielded a positive mean difference and achieved statistical significance ($\alpha = 0.05$). Where multiple variants were tested, $p$-values are descriptive; stage-wise increments are also reported with Bonferroni correction (Appendix~\ref{sec:appendix-ablation-stages}).

\paragraph{Drift Control}
Every stage was compared with a recomputed reference run rather than with a previously recorded value, so that changes in library versions, hardware or random state could not be mistaken for improvements, and these recomputations agree with the recorded values to six decimal places.

\paragraph{Mechanistic Transparency over Metric Optimization}
When two hyperparameter configurations yielded statistically indistinguishable performance on the tuning set, the variant with the clearer theoretical foundation was retained. For instance, the relative context rule $W_\gamma = 64L$, which ties the contrast context to the window length $L$ and hence to the dominant period of each series, was selected over an absolute window of $32{,}000$ samples. Similarly, settings that raised overall mean metrics at the cost of severe performance degradation on a substantial minority of series were systematically rejected. Full hardware and software specifications are detailed in Appendix~\ref{sec:appendix-hardware}.


\section{Experimental Results and Analysis}
\label{sec:results}

This section compares \LEAP{} with the state of the art on both tracks and isolates the mechanisms behind its performance through ablation, sensitivity and stratified analyses. All paired tests are computed on per-series values rather than on aggregates: baseline comparisons use three-seed means on the held-out series, and the sensitivity analyses use the tuning set.

\subsection{Performance on TSB-AD-M}
\label{sec:results-overall}

Table~\ref{tab:overall} compares \LEAP{} with $21$ baseline anomaly detectors on the $180$ held-out multivariate series. Sixteen of the twenty-one baselines were re-run locally from their released source code using the same TSB-AD evaluation protocol, file list, and three random seeds ($2024$, $2025$, and $2026$). The table reports published values alongside locally measured means and sample standard deviations where available. \LEAP{} was evaluated under the same protocol, with its per-series results archived alongside those of the locally evaluated baselines.

\begin{table}[htbp]
\centering
\small
\caption{Mean VUS-PR on the TSB-AD-M held-out set (180 series). Published and locally measured results are reported where available; measured values are mean $\pm$ sample standard deviation over three seeds. \emph{Source}: reference from which the published value is taken; for rows without a published value, the reference describing the method. Daggers ($\dagger$) designate standalone published detectors outside the standard TSB-AD benchmark suite.}
\label{tab:overall}
\begin{tabular}{llccl}
\toprule
Method & Family & Published & Measured (this work) & Source \\
\midrule
\textbf{\LEAP{} (this work)} & Masked $+$ repr. & --- & $\mathbf{0.5179 \pm 0.0022}$ & this work \\
\midrule
VACE$^\dagger$ & Representation & 0.48 & --- & \cite{cencillo2026vace} \\
PaAno$^\dagger$ & Representation & 0.43 & $0.4161 \pm 0.0064$ & \cite{park2026paano} \\
PaAno$+^\dagger$ & Representation & 0.42 & --- & \cite{zhu2026paanoplus} \\
PaAno$_{\mathrm{PAI}}$ & Representation & --- & $0.4045 \pm 0.0024$ & \cite{zhang2026pai} \\
KAN-AD$^\dagger$ & Forecasting & 0.41 & $0.3241 \pm 0.0007$ & \cite{zhou2025kanad} \\
xLSTMAD-R$^\dagger$ & Reconstruction & 0.37 & $0.3881 \pm 0.0011$ & \cite{faber2025xlstmad} \\
MMPAD$^\dagger$ & Discord & 0.35 & --- & \cite{yeh2026mpbenchmark} \\
xLSTMAD-F$^\dagger$ & Forecasting & 0.35 & $0.3490 \pm 0.0019$ & \cite{faber2025xlstmad} \\
CrossAD$^\dagger$ & Reconstruction & 0.33 & $0.3355 \pm 0.0014$ & \cite{li2025crossad} \\
CNN & Forecasting & 0.31 & $0.3538 \pm 0.0007$ & \cite{liu2024tsbad} \\
OmniAnomaly & Reconstruction & 0.31 & $0.3152 \pm 0.0001$ & \cite{cencillo2026vace} \\
DADA$^\dagger$ & Reconstruction & 0.31 & --- & \cite{shentu2025dada} \\
PCA & Statistical & 0.31 & $0.2691 \pm 0.0000$ & \cite{liu2024tsbad} \\
USAD & Reconstruction & 0.30 & $0.4028 \pm 0.0019$ & \cite{cencillo2026vace} \\
CATCH$^\dagger$ & Reconstruction & 0.30 & $0.3303 \pm 0.0006$ & \cite{wu2025catch} \\
IForest & Statistical & 0.20 & $0.2469 \pm 0.0024$ & \cite{cencillo2026vace} \\
TimesNet & Reconstruction & 0.19 & $0.3874 \pm 0.0022$ & \cite{park2026paano} \\
TranAD & Transformer & 0.18 & --- & \cite{zhu2026paanoplus} \\
CARLA$^\dagger$ & Contrastive SSL & --- & $0.1351 \pm 0.0204$ & \cite{darban2024carla} \\
Anomaly Transf.\ & Transformer & 0.12 & $0.1147 \pm 0.0018$ & \cite{park2026paano} \\
DCdetector & Transformer & 0.09 & $0.0962 \pm 0.0005$ & \cite{cencillo2026vace} \\
\bottomrule
\end{tabular}
\end{table}

The comparison distinguishes published results from standardized local evaluations. Five baselines (VACE, PaAno$+$, MMPAD, DADA, and TranAD) are represented only by published values because their source code was unavailable. Among the $16$ locally evaluated baselines, substantial discrepancies from published scores occur in both directions: KAN-AD drops from approximately $0.41$ to $0.3241 \pm 0.0007$, while USAD and TimesNet increase from $0.30$ and $0.19$ to $0.4028 \pm 0.0019$ and $0.3874 \pm 0.0022$, respectively. PaAno changes less, from $0.43$ to $0.4161 \pm 0.0064$. These discrepancies highlight the limitations of ranking methods solely by published scores.

\LEAP{} achieves the highest locally measured VUS-PR ($0.5179 \pm 0.0022$; Table~\ref{tab:overall}), outperforming the strongest measured baseline, PaAno ($0.4161$), by $24.5\%$. It also exceeds published PaAno ($0.43$) by $20.4\%$ and the best published baseline, VACE ($0.48$), by $7.9\%$, although these cross-protocol comparisons should be interpreted cautiously. The absolute margins over PaAno$_{\mathrm{PAI}}$, USAD, and measured KAN-AD are $+0.1134$, $+0.1151$, and $+0.1938$, respectively. The close agreement with the single-seed ablation result ($0.5171$; Table~\ref{tab:ablation-m}) is consistent with the low variability across seeds. Baselines represented only by published results are excluded from subsequent paired statistical analyses.

\begin{figure}[!htbp]
\centering
\includegraphics[width=0.65\linewidth]{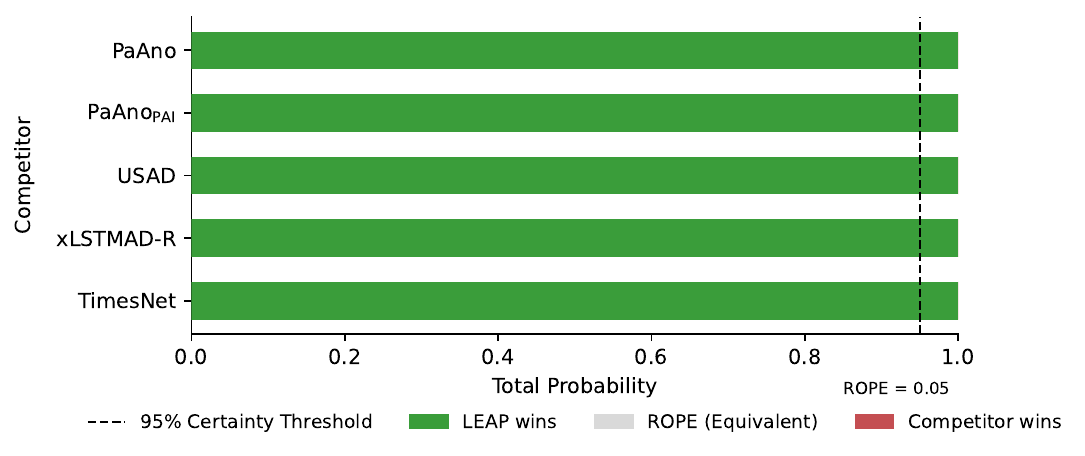}
\caption{Bayesian signed-rank test comparing
\LEAP{} against each of the five measured competitors of
Table~\ref{tab:overall}, with a region of practical equivalence of $\pm0.05$. Green
is the posterior probability that \LEAP{} is superior, gray the probability
of practical equivalence, and red the probability that the competitor is
superior.}
\label{fig:bayes-signedrank}
\end{figure}

Aggregate averages can obscure substantial differences across datasets: per-series VUS-PR scores exhibit a strongly bimodal distribution, with median scores $\leq 0.16$ for all six methods on GHL but $\geq 0.71$ on Exathlon and TAO. This motivates examining per-series distributions (Figure~\ref{fig:overall-boxplot}) and paired statistical tests, alongside the five-stage ablation in Appendix~\ref{sec:appendix-ablation}. A Bayesian signed-rank test~\cite{benavoli14} compares \LEAP{} with the five strongest locally measured baselines (PaAno, PaAno$_{\mathrm{PAI}}$, USAD, xLSTMAD-R, and TimesNet), using three-seed mean VUS-PR scores across the $180$ paired series and a ROPE of $\pm0.05$. As shown in Figure~\ref{fig:bayes-signedrank}, \LEAP{} achieves a posterior probability of superiority above $99.9\%$ in all five comparisons, with negligible posterior probability of any competitor outperforming it.

The nine-metric comparison in Table~\ref{tab:ninemetric} further supports this result. \LEAP{} achieves the highest score on eight of nine metrics: AUC-ROC ($0.8410$), AUC-PR ($0.4794$), VUS-ROC ($0.8535$), VUS-PR ($0.5179$), Standard-F1 ($0.5427$), Event-based-F1 ($0.6285$), R-based-F1 ($0.5462$), and Affiliation-F ($0.8821$). The only exception is PA-F1, where PaAno$_{\mathrm{PAI}}$ reaches $0.7684$ versus $0.7290$ for \LEAP{} ($0.0394$ higher). This divergence reflects the point-wise versus range-aware trade-off discussed in Section~\ref{sec:bg-benchmarks}: consistent with Section~\ref{sec:exp-metrics} and Kim \cite{kim2022rigorous}, PA-F1 credits an entire ground-truth anomaly segment when a single anomalous point is detected, favoring PaAno$_{\mathrm{PAI}}$'s point-level calibration. PA-F1 should therefore be interpreted as a diagnostic indicator of metric behavior rather than a definitive measure of detector quality. TimesNet shows a related discrepancy: although it ranks last among the six methods on threshold-free and point-wise metrics (AUC-ROC, AUC-PR, VUS-ROC, VUS-PR, and Standard-F1), it ranks third on PA-F1 ($0.6644$) and fourth on Event-based-F1, R-based-F1, and Affiliation-F.

\begin{table}[htbp]
\setlength{\tabcolsep}{3pt}
\centering
\small
\caption{Nine-metric evaluation on the TSB-AD-M held-out set (180 series) for the top five measured competitors. Values are mean $\pm$ sample standard deviation over three seeds; the best value in each row is bold.}
\label{tab:ninemetric}
\begin{tabular}{lcccccc}
\toprule
Metric & PaAno & PaAno$_{\mathrm{PAI}}$ & USAD & xLSTMAD-R & TimesNet & \LEAP{} \\
\midrule
AUC-ROC & $0.7617 \pm 0.0048$ & $0.7833 \pm 0.0008$ & $0.7365 \pm 0.0005$ & $0.7387 \pm 0.0016$ & $0.6681 \pm 0.0016$ & $\mathbf{0.8410 \pm 0.0028}$ \\
AUC-PR & $0.3763 \pm 0.0056$ & $0.3979 \pm 0.0008$ & $0.3617 \pm 0.0008$ & $0.3413 \pm 0.0017$ & $0.3358 \pm 0.0016$ & $\mathbf{0.4794 \pm 0.0031}$ \\
VUS-ROC & $0.7887 \pm 0.0044$ & $0.8057 \pm 0.0007$ & $0.7659 \pm 0.0012$ & $0.7773 \pm 0.0007$ & $0.7090 \pm 0.0023$ & $\mathbf{0.8535 \pm 0.0036}$ \\
VUS-PR & $0.4161 \pm 0.0064$ & $0.4045 \pm 0.0024$ & $0.4028 \pm 0.0019$ & $0.3881 \pm 0.0011$ & $0.3874 \pm 0.0022$ & $\mathbf{0.5179 \pm 0.0022}$ \\
Standard-F1 & $0.4271 \pm 0.0046$ & $0.4366 \pm 0.0020$ & $0.4333 \pm 0.0007$ & $0.4221 \pm 0.0007$ & $0.4098 \pm 0.0017$ & $\mathbf{0.5427 \pm 0.0017}$ \\
PA-F1 & $0.6521 \pm 0.0047$ & $\mathbf{0.7684 \pm 0.0015}$ & $0.6248 \pm 0.0033$ & $0.6216 \pm 0.0057$ & $0.6644 \pm 0.0015$ & $0.7290 \pm 0.0020$ \\
Event-based-F1 & $0.5124 \pm 0.0061$ & $0.6064 \pm 0.0018$ & $0.5198 \pm 0.0015$ & $0.5339 \pm 0.0056$ & $0.5323 \pm 0.0043$ & $\mathbf{0.6285 \pm 0.0045}$ \\
R-based-F1 & $0.4063 \pm 0.0124$ & $0.3836 \pm 0.0032$ & $0.4229 \pm 0.0011$ & $0.3835 \pm 0.0004$ & $0.4031 \pm 0.0036$ & $\mathbf{0.5462 \pm 0.0024}$ \\
Affiliation-F & $0.8516 \pm 0.0018$ & $0.8659 \pm 0.0013$ & $0.8371 \pm 0.0003$ & $0.8332 \pm 0.0012$ & $0.8452 \pm 0.0034$ & $\mathbf{0.8821 \pm 0.0020}$ \\
\bottomrule
\end{tabular}
\end{table}

Figure~\ref{fig:overall-boxplot} shows the per-series VUS-PR distributions across ten source datasets, covering $169$ of the $180$ series. \LEAP{} obtains the highest mean on seven datasets (SVDB, SMAP, Exathlon, GHL, MSL, MITDB, and CATSv2), while individual baselines perform better on SMD, TAO, and OPPORTUNITY. The largest baseline advantage occurs on OPPORTUNITY, where USAD exceeds \LEAP{} by $0.2808$ in mean VUS-PR. These dataset-specific differences show that the overall ranking reflects performance across heterogeneous domains rather than uniform superiority on every dataset. Appendix~\ref{sec:appendix-domain-breakdown} extends this analysis with mean VUS-PR by source dataset and application domain (Tables~\ref{tab:breakdown-m} and~\ref{tab:breakdown-u}), by dominant anomaly type (Tables~\ref{tab:breakdown-type-m} and~\ref{tab:breakdown-type-u}), and by series characteristics (Tables~\ref{tab:breakdown-char-m} and~\ref{tab:breakdown-char-u}).

\begin{figure}[!htbp]
\centering
\includegraphics[width=0.9\linewidth]{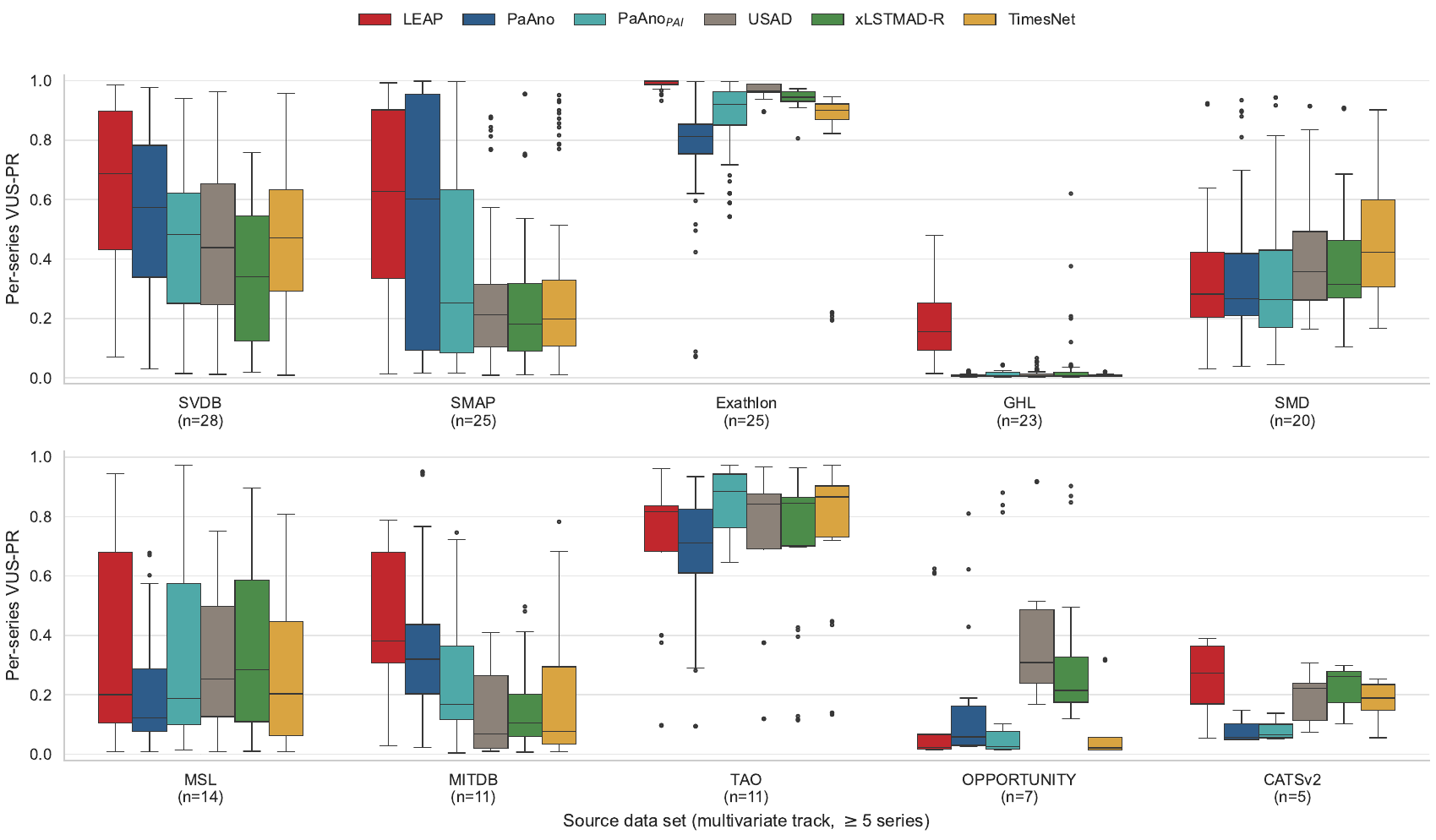}

\caption{Per-series VUS-PR distributions on TSB-AD-M eval set for \LEAP{} and the top five measured competitors (Table~\ref{tab:overall}), pooled across three random seeds and grouped by source dataset (restricted to $10$ datasets with $\ge 5$ series; $169$ of $180$ series total represented). Boxes indicate median and IQR; whiskers extend to $1.5\times\text{IQR}$, with outliers plotted individually.}

\label{fig:overall-boxplot}
\end{figure}

\subsection{Sensitivity to Fusion Coefficients}
\label{sec:results-sensitivity}

\begin{figure}[!htbp]
\centering
\includegraphics[width=0.8\linewidth]{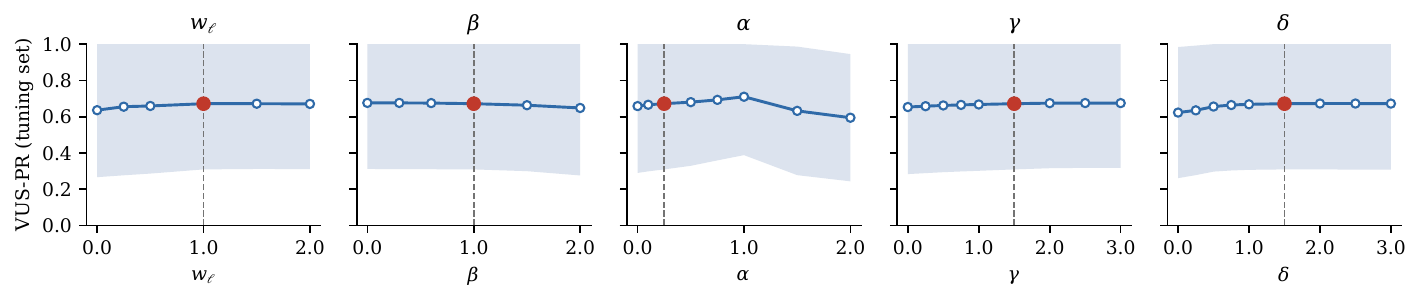}
\caption{VUS-PR sensitivity to the five fusion coefficients on the $20$-series tuning set. Shaded regions denote one standard deviation across benchmark series.}
\label{fig:sensitivity}
\end{figure}

To assess generalization across heterogeneous time series, we swept $w_\ell$, $\beta$, $\alpha$, $\gamma$, and $\delta$ over $20$ tuning series from $11$ dataset families. Figure~\ref{fig:sensitivity} and Table~\ref{tab:sensitivity-coeffs} show that cross-series heterogeneity outweighs hyperparameter sensitivity: deployed VUS-PR ranges from $0.049$ to $0.999$ (standard deviation $=0.36$), while selecting each coefficient's grid optimum improves the mean by at most $0.0384$. The deployed values are therefore close to optimal for four coefficients, with $\alpha$ showing greater sensitivity.

\begin{table}[htbp]
\centering
\small
\caption{Sensitivity of the fusion coefficients on the $20$-series tuning set. Each parameter is varied independently while the remaining parameters are held at their deployed values ($w_\ell=\beta=1.0$, $\alpha=0.25$, $\gamma=\delta=1.5$). Values $\alpha > 1$ exceed the recommended range of Eq.~\eqref{eq:lof} and are included only to illustrate the risk of near-zero denominators.}
\label{tab:sensitivity-coeffs}
\begin{tabular}{lcccll}
\toprule
Coefficient & Grid tested & Points & Deployed value & VUS-PR (deployed) & Grid optimum \\
\midrule
$w_\ell$ & $0$ to $2.0$ & $6$ & $1.0$  & $0.6717$ & $0.6717$ at $1.5$ ($\Delta < 0.0001$) \\
$\beta$  & $0$ to $2.0$ & $6$ & $1.0$  & $0.6717$ & $0.6761$ at $0$ ($\Delta = 0.0044$) \\
$\alpha$ & $0$ to $2.0$ & $8$ & $0.25$ & $0.6717$ & $0.7101$ at $1.0$ ($\Delta = 0.0384$) \\
$\gamma$ & $0$ to $3.0$ & $9$ & $1.5$  & $0.6717$ & $0.6753$ at $2.5$ ($\Delta = 0.0036$) \\
$\delta$ & $0$ to $3.0$ & $9$ & $1.5$  & $0.6717$ & $0.6726$ at $2.0$ ($\Delta = 0.0009$) \\
\bottomrule
\end{tabular}
\end{table}

All evaluations were restricted to the tuning set, with parameter selection verified through leave-one-series-out (LOSO) cross-validation rather than direct peak selection. The apparent gains for $w_\ell$ and $\beta$ disappeared when a single favorable series was excluded, while $\alpha$ required excluding two. These influential series (\texttt{011\_MSL}, \texttt{062\_SMD}, and \texttt{131\_OPPORTUNITY}) highlight cross-series heterogeneity. Although $\alpha=1.0$ achieves the highest aggregate VUS-PR ($0.7101$), its gains are concentrated in OPPORTUNITY and MITDB, while degrading MSL, SMD, and GHL. This optimum also lies near the denominator instability described in Section~\ref{sec:leap-stage3}. The sweep extends to $2.0$ with a numerical safeguard, although $[0,1]$ remains the recommended range. Thus, the deployed configuration retains $w_\ell=\beta=1.0$ and $\alpha=0.25$, prioritizing robustness over aggregate optima.

In contrast, the local contrast coefficient $\gamma$ and channel-pool novelty weight $\delta$ were the only parameters showing consistent gains. Both improved over their zero-value baselines across all $20$ LOSO folds, with mean VUS-PR gains of $+0.018$ ($10/3/7$ W/L/T, $p$-value $=0.049$) and $+0.049$ ($13/4/3$, $p$-value $=0.016$), respectively. Their deployed value of $1.5$ is also close to the grid optima, differing by only $0.0036$ for $\gamma$ and $0.0009$ for $\delta$, suggesting broad performance plateaus rather than isolated optima (Figure~\ref{fig:sensitivity}).

The analysis also examined combination rules and five backbone parameters (Appendix~\ref{app:sensitivity}). No alternative combination rule demonstrated a consistent or statistically significant advantage over the deployed \texttt{add} and \texttt{softmax} operators. Among backbone parameters, only patch length significantly affected performance: $p=8$ ($\Delta=-0.0454$, $p$-value $=0.003$) and $p=32$ ($\Delta=-0.0347$, $p$-value $=0.014$) both degraded performance, whereas depth, width, masking groups, and mask block size showed no significant effects ($p$-values $\geq 0.225$). These findings suggest that \LEAP{}'s gains stem primarily from its scoring and fusion mechanisms rather than backbone capacity tuning.

\subsection{Results on TSB-AD-U}
\label{sec:results-u}

Table~\ref{tab:overall-u} compares \LEAP{} with $23$ baseline methods on the $350$ held-out univariate series of TSB-AD-U, following the protocol described in Section~\ref{sec:results-overall}.

\begin{table}[htbp]
\centering
\small
\caption{Mean VUS-PR performance on TSB-AD-U. \emph{Published} columns report published values (ranges indicate cited intervals), while \emph{Measured} columns report local mean $\pm$ standard deviation, with dashes denoting unavailable entries. Method \emph{Family} classifications extend Table~\ref{tab:families} with \emph{Foundation model} (TSPulse, MOMENT, TimesFM) and \emph{Shape clustering} (KShapeAD). Daggers ($\dagger$) designate standalone published detectors outside the standard TSB-AD benchmark suite. \emph{Source}: reference from which the published value is taken; for rows
without a published value, the reference describing the method.}
\label{tab:overall-u}
\begin{tabular}{llccl}
\toprule
Method & Family & Published & Measured (this work) & Source \\
\midrule
\textbf{\LEAP{} (this work)} & Masked $+$ repr. & --- & $\mathbf{0.6194 \pm 0.0016}$ & this work \\
\midrule
PaAno$_{\mathrm{PAI}}$ & Representation & 0.58 & $0.5839 \pm 0.0016$ & \cite{zhang2026pai} \\
PaAno$+^\dagger$ & Representation & 0.55 & --- & \cite{zhu2026paanoplus} \\
TSPulse$^\dagger$ & Foundation model & 0.54 & --- & \cite{ekambaram2025tspulse} \\
PaAno$^\dagger$ & Representation & 0.51--0.53 & $0.5165 \pm 0.0025$ & \cite{park2026paano} \\
CrossAD$^\dagger$ & Reconstruction & 0.45 & --- & \cite{li2025crossad} \\
PatchTST & Forecasting & --- & $0.4569 \pm 0.0025$ & \cite{liu2024tsbad} \\
Sub-PCA & Statistical & 0.42 & $0.4173 \pm 0.0000$ & \cite{yeh2026mpbenchmark} \\
KAN-AD$^\dagger$ & Forecasting & 0.40--0.43 & $0.3695 \pm 0.0125$ & \cite{zhou2025kanad} \\
USAD & Reconstruction & --- & $0.3959 \pm 0.0017$ & \cite{liu2024tsbad} \\
FITS & Forecasting & --- & $0.3902 \pm 0.0070$ & \cite{liu2024tsbad} \\
CNN & Forecasting & --- & $0.3823 \pm 0.0068$ & \cite{liu2024tsbad} \\
KShapeAD & Shape clustering & --- & $0.3750 \pm 0.0063$ & \cite{liu2024tsbad} \\
MOMENT-FT$^\dagger$ & Foundation model & --- & $0.3707 \pm 0.0012$ & \cite{goswami2024moment} \\
TimesNet & Reconstruction & --- & $0.3707 \pm 0.0044$ & \cite{liu2024tsbad} \\
MOMENT-ZS$^\dagger$ & Foundation model & --- & $0.3662 \pm 0.0009$ & \cite{goswami2024moment} \\
iTransformer & Forecasting & --- & $0.3585 \pm 0.0003$ & \cite{liu2024tsbad} \\
OFA & Forecasting & --- & $0.3581 \pm 0.0029$ & \cite{liu2024tsbad} \\
DLinear & Forecasting & --- & $0.3493 \pm 0.0017$ & \cite{liu2024tsbad} \\
NLinear & Forecasting & --- & $0.3333 \pm 0.0015$ & \cite{liu2024tsbad} \\
TimesFM$^\dagger$ & Foundation model & --- & $0.2964 \pm 0.0000$ & \cite{liu2024tsbad} \\
CARLA$^\dagger$ & Contrastive SSL & --- & $0.1136 \pm 0.0012$ & \cite{darban2024carla} \\
Anomaly Transf.\ & Transformer & --- & $0.1035 \pm 0.0049$ & \cite{liu2024tsbad} \\
DCdetector & Transformer & --- & $0.0884 \pm 0.0006$ & \cite{liu2024tsbad} \\
\bottomrule
\end{tabular}
\end{table}

\LEAP{} achieves the best overall performance, with a measured VUS-PR of $0.6194 \pm 0.0016$, outperforming all $20$ locally evaluated baselines, including PaAno$_{\mathrm{PAI}}$ ($0.5839$, $\Delta = +0.0355$, $+6.1\%$), and PaAno$+$ ($0.55$). The advantage reaches $0.1029$ over the standalone PaAno implementation ($0.5165$). Conversely, PaAno$_{\mathrm{PAI}}$ substantially outperforms standalone PaAno despite sharing the same method family, increasing VUS-PR by more than $0.06$ under the same evaluation protocol and highlighting the impact of benchmark-specific variants and scoring configurations.

The KAN-AD discrepancy reported for TSB-AD-M in Section~\ref{sec:results-overall} is also observed on TSB-AD-U: its measured score ($0.3695 \pm 0.0125$) is approximately $0.03$ below the lower bound of the published range ($0.40$--$0.43$), underscoring the sensitivity of reported performance to evaluation implementation. Performance also varies across algorithmic families. \emph{Representation}-based methods lead most baseline rankings, while PAI post-processing lifts PaAno to $0.58$ VUS-PR, above the foundation models TSPulse ($0.54$), MOMENT and TimesFM ($0.30$--$0.37$). The \emph{Forecasting} family ranges from PatchTST ($0.4569$) to NLinear ($0.3333$), limiting the predictive value of family-level taxonomy. \emph{Reconstruction} and \emph{Statistical/Shape-based} methods form a narrower middle tier ($0.37$--$0.42$), while the contrastive CARLA ($0.1136$) and pure \emph{Transformer} baselines remain at the bottom, consistent with the multivariate results.

\begin{figure}[!htbp]
\centering
\includegraphics[width=0.55\linewidth]{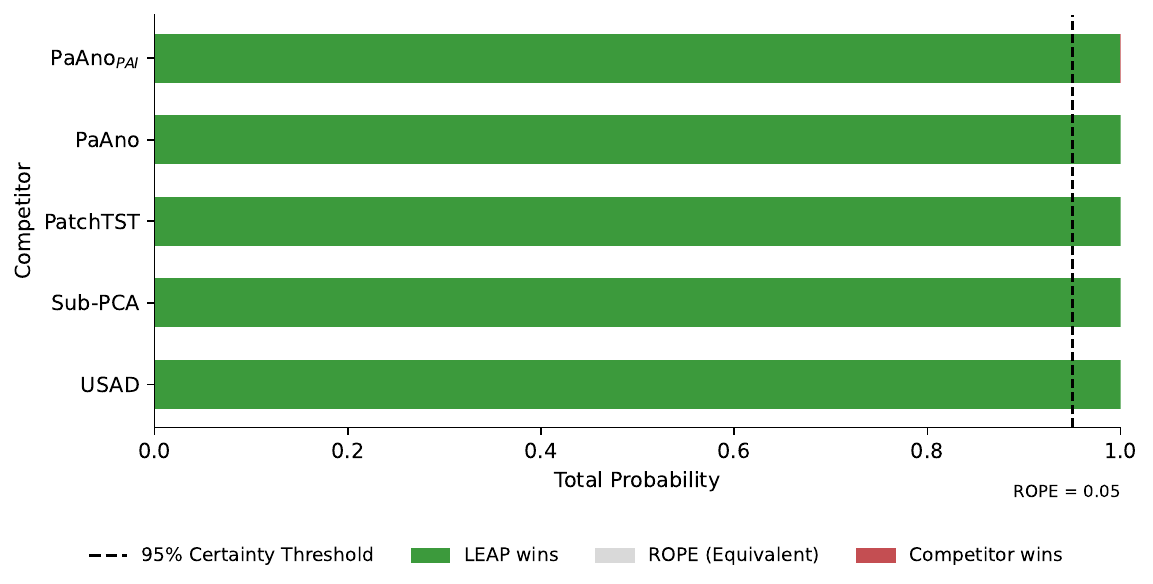}
\caption{Bayesian signed-rank test comparing \LEAP{} against the top five measured univariate competitors from Table~\ref{tab:overall-u}.}
\label{fig:bayes-signedrank-u}
\end{figure}

Following Section~\ref{sec:results-overall}, Figure~\ref{fig:bayes-signedrank-u} summarizes the Bayesian signed-rank comparisons between \LEAP{} and the five strongest measured baselines across the $350$ paired series ($\text{ROPE} = \pm 0.05$ VUS-PR). \LEAP{} achieves a posterior probability of superiority above $99.9\%$ against all five competitors (PaAno$_{\mathrm{PAI}}$, PaAno, PatchTST, Sub-PCA, and USAD), with negligible posterior mass assigned to practical equivalence or competitor superiority. Against PaAno$_{\mathrm{PAI}}$, \LEAP{} wins on $202$ of $350$ series ($57.7\%$), with a mean VUS-PR advantage of $+0.0355$.

Table~\ref{tab:ninemetric-u} extends the comparison to nine evaluation metrics for the five strongest locally measured baselines, following the multivariate analysis in Table~\ref{tab:ninemetric}. \LEAP{} leads in five metrics (AUC-ROC, VUS-ROC, VUS-PR, Standard-F1, and R-based-F1), while PaAno$_{\mathrm{PAI}}$ leads in the remaining four: AUC-PR ($0.5506$ vs.\ $0.5409$), PA-F1 ($0.8375$ vs.\ $0.7447$), Event-based-F1 ($0.7532$ vs.\ $0.6860$), and Affiliation-F ($0.9278$ vs.\ $0.9194$). These results indicate that \LEAP{}'s advantage in VUS-PR does not extend uniformly across evaluation metrics.

\begin{table}[htbp]
\centering
\small
\caption{Nine-metric evaluation on the TSB-AD-U held-out set (350 series). Values are mean $\pm$ sample standard deviation across the three seeds; Sub-PCA is deterministic and therefore has zero spread. The six methods are evaluated locally, and the best value in each row is bold.}
\label{tab:ninemetric-u}
\begin{tabular}{lcccccc}
\toprule
Metric & PaAno$_{\mathrm{PAI}}$ & PaAno & PatchTST & Sub-PCA & USAD & \LEAP{} \\
\midrule
AUC-ROC & $0.9022{\pm}.0005$ & $0.8663{\pm}.0005$ & $0.7692{\pm}.0020$ & $0.7143{\pm}.0000$ & $0.7027{\pm}.0021$ & $\mathbf{0.9128{\pm}.0005}$ \\
AUC-PR & $\mathbf{0.5506{\pm}.0015}$ & $0.4695{\pm}.0025$ & $0.3676{\pm}.0024$ & $0.3777{\pm}.0000$ & $0.3631{\pm}.0013$ & $0.5409{\pm}.0022$ \\
VUS-ROC & $0.9280{\pm}.0002$ & $0.8837{\pm}.0007$ & $0.8235{\pm}.0004$ & $0.7585{\pm}.0000$ & $0.7493{\pm}.0016$ & $\mathbf{0.9295{\pm}.0008}$ \\
VUS-PR & $0.5839{\pm}.0016$ & $0.5165{\pm}.0025$ & $0.4569{\pm}.0025$ & $0.4173{\pm}.0000$ & $0.3959{\pm}.0017$ & $\mathbf{0.6194{\pm}.0016}$ \\
Standard-F1 & $0.5789{\pm}.0027$ & $0.5166{\pm}.0013$ & $0.4330{\pm}.0025$ & $0.4168{\pm}.0000$ & $0.4029{\pm}.0014$ & $\mathbf{0.5849{\pm}.0004}$ \\
PA-F1 & $\mathbf{0.8375{\pm}.0019}$ & $0.6730{\pm}.0053$ & $0.6822{\pm}.0023$ & $0.5719{\pm}.0000$ & $0.5431{\pm}.0019$ & $0.7447{\pm}.0005$ \\
Event-based-F1 & $\mathbf{0.7532{\pm}.0029}$ & $0.5847{\pm}.0034$ & $0.5575{\pm}.0038$ & $0.5039{\pm}.0000$ & $0.4660{\pm}.0010$ & $0.6860{\pm}.0012$ \\
R-based-F1 & $0.5121{\pm}.0025$ & $0.4879{\pm}.0031$ & $0.4187{\pm}.0005$ & $0.4097{\pm}.0000$ & $0.3978{\pm}.0021$ & $\mathbf{0.5500{\pm}.0024}$ \\
Affiliation-F & $\mathbf{0.9278{\pm}.0008}$ & $0.8959{\pm}.0014$ & $0.8960{\pm}.0011$ & $0.8529{\pm}.0000$ & $0.8343{\pm}.0017$ & $0.9194{\pm}.0011$ \\
\bottomrule
\end{tabular}
\end{table}

The comparison with PaAno$_{\mathrm{PAI}}$ highlights the importance of amplitude information. PAI restores magnitude sensitivity through post-hoc corrections \cite{zhang2026pai}, raising PaAno to $0.5839$ VUS-PR, whereas \LEAP{} preserves amplitude natively through unnormalized masked prediction and its local-moment module (Section~\ref{sec:leap-moments}), without requiring post-hoc correction. The remaining $15$ locally evaluated baselines in Table~\ref{tab:overall-u} fall outside the $\pm 0.05$ ROPE relative to \LEAP{}, so further pairwise tests are unnecessary.

\begin{figure}[!htbp]
\centering
\includegraphics[width=0.8\linewidth]{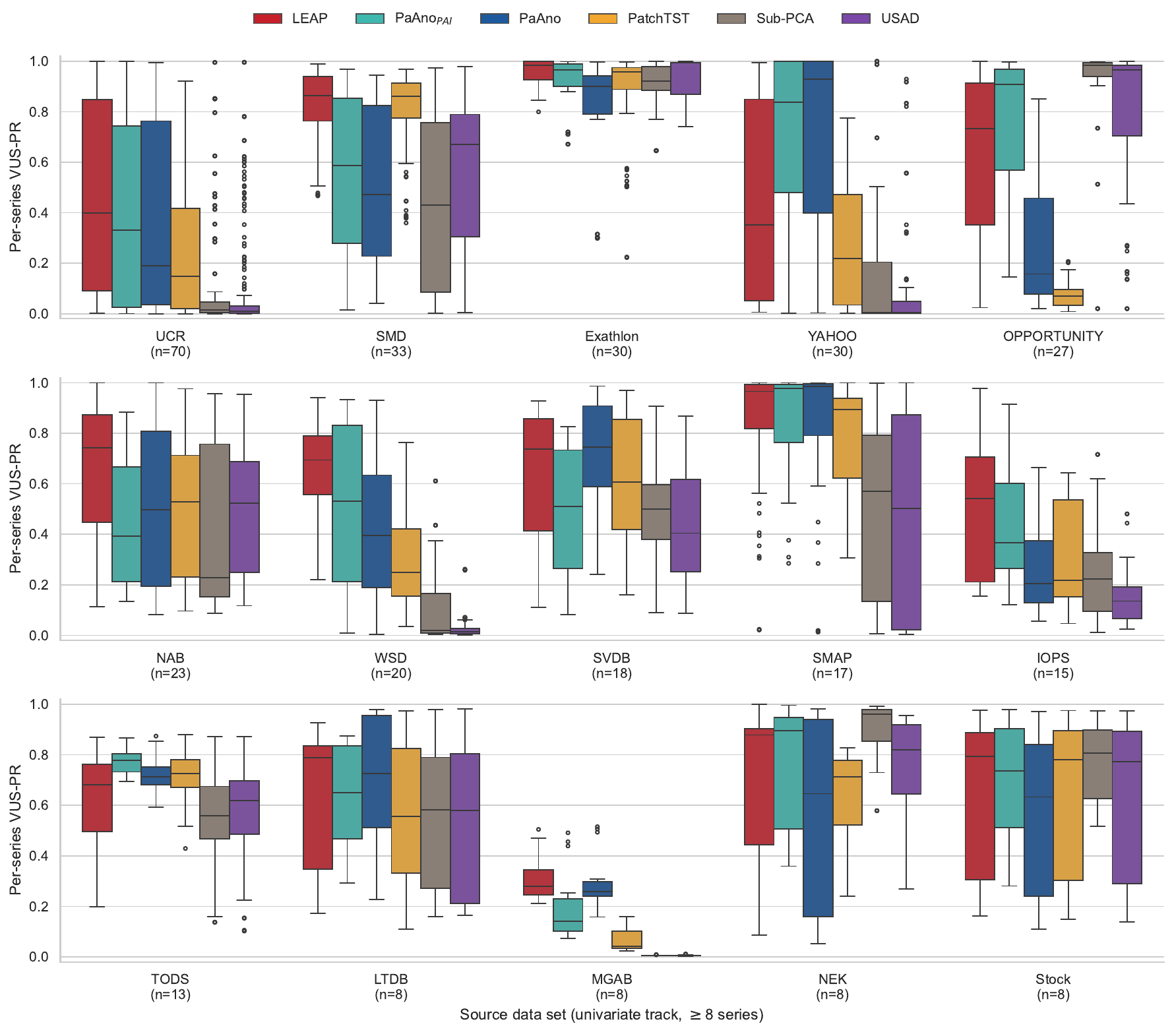}
\caption{Per-series VUS-PR distributions on the TSB-AD-U held-out set for \LEAP{} and the five competitors, grouped by source dataset. Results cover the fifteen datasets with at least eight series ($328$ of $350$ series); boxes show the interquartile range and median, whiskers extend to $1.5\times$ the interquartile range, and points beyond the whiskers are shown individually.}
\label{fig:overall-boxplot-u}
\end{figure}

Figure~\ref{fig:overall-boxplot-u} shows that performance remains dataset-dependent, although TSB-AD-U exhibits more continuous variation than TSB-AD-M. \LEAP{} is particularly competitive on UCR, where it achieves the highest mean VUS-PR ($0.459$) compared with $0.402$ for PaAno$_{\mathrm{PAI}}$, and ranks first on seven of the fifteen datasets. YAHOO remains a clear weakness, with a median of $0.352$ versus $0.929$ for PaAno, whereas on OPPORTUNITY \LEAP{} reaches a median of $0.734$. Exathlon shows consistently high performance across methods, with \LEAP{} reaching a median of $0.984$ (mean $0.9626$; Table~\ref{tab:breakdown-u}).

\subsection{Qualitative Analysis}
\label{sec:results-qualitative}

\begin{figure}[!htpb]
  \centering
  \includegraphics[width=0.9\linewidth]{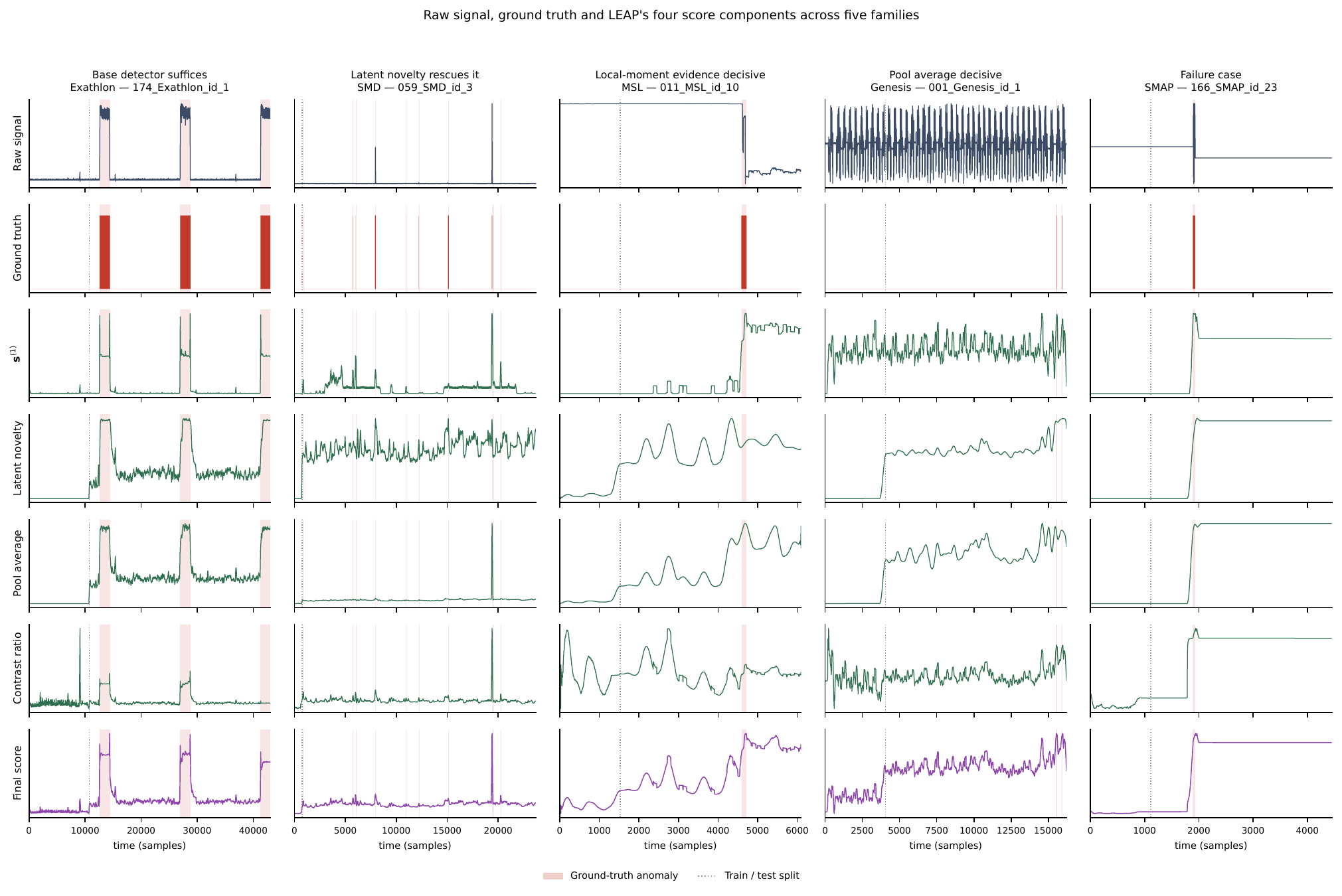}
    \caption{Raw signal, ground-truth annotations, \LEAP{}'s four score components
($s^{(1)}$, which fuses masked prediction and local-moment evidence; latent
novelty; per-channel pooling; local contrast) and the fused final score for five
representative series. Signals correspond to the continuous channel with the largest standardized mean shift between anomalous and normal regions, avoiding maximum-variance selection that can favor binary telemetry flags.}
  \label{fig:qualitative}
\end{figure}

\begin{figure}[!htpb]
  \centering
  \includegraphics[width=0.5\linewidth]{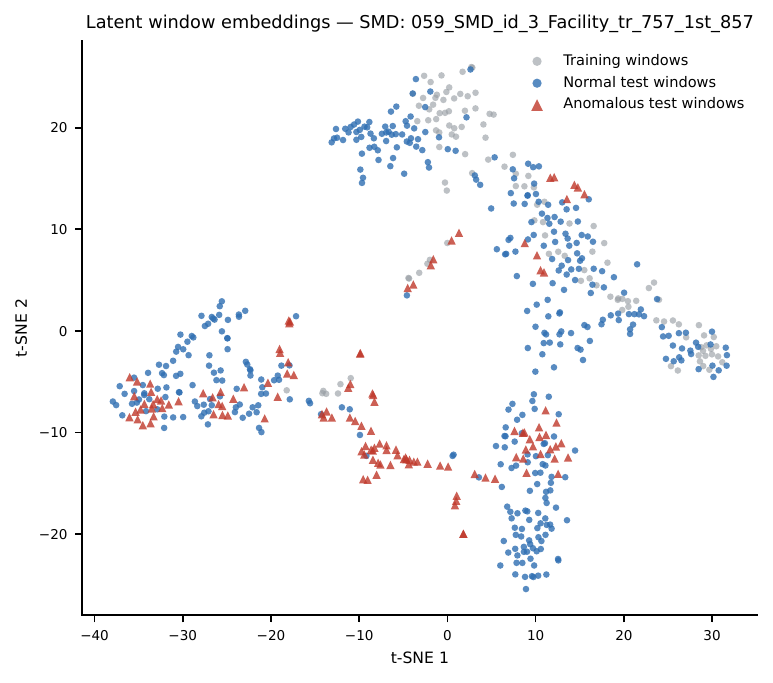}
  \caption{Two-dimensional t-SNE projection of per-window latent embeddings (Eq.~\eqref{eq:embed}) for \texttt{059\_SMD\_id\_3}, showing training, normal test, and anomalous test windows.}
  \label{fig:embedding}
\end{figure}

To quantify the behaviors in Figure~\ref{fig:qualitative}, each evidence component was evaluated \emph{in isolation} on its corresponding series using the standard VUS-PR protocol. Although some components are not standalone detectors (e.g., contrast depends on $s^{(1)}$), this isolates their contributions before score fusion:

\begin{itemize}
\item \textbf{Base detector sufficiency in Exathlon (\texttt{174\_Exathlon\_id\_1}).}
Masked prediction already achieves $0.9998$ VUS-PR ($s^{(1)}=0.9848$), as the high-amplitude bursts closely match the anomaly patterns targeted by masked-patch prediction. Subsequent layers reinforce this response, illustrating the regime where anomalies are clear departures from normal dynamics.

\item \textbf{Partial recovery via latent novelty in SMD (\texttt{059\_SMD\_id\_3}).}
Masked prediction is weak ($0.1486$ VUS-PR) because the brief, low-amplitude anomalies produce little prediction error. Latent $k$-nearest-neighbor novelty improves this to $0.2442$, while local contrast reaches $0.3297$, yielding $0.2669$ after fusion. The recovery remains partial because some anomalous windows remain embedded in the normal latent manifold. The space can be seen geometrically in Figure~\ref{fig:embedding}.

\item \textbf{Decisive local-moment evidence in MSL (\texttt{011\_MSL\_id\_10}).}
Local-moment evidence reaches $0.8790$ VUS-PR versus only $0.0936$ for masked prediction, as $47$ of $55$ channels are near-constant command flags. It therefore captures abrupt state transitions missed by prediction, while the shape-based terms contribute little. This illustrates the failure mode motivating local-moment evidence.

\item \textbf{Synergy through per-channel pooling in Genesis (\texttt{001\_Genesis\_id\_1}).}
The components of $s^{(1)}$ conflict (masked prediction: $0.7004$; level: $0.0580$), reducing their combined score to $0.3966$. Although pooling is weak in isolation ($0.1102$), its fusion with local contrast raises the final score to $0.7286$. This suggests that complementary ranking errors can be corrected through additive fusion.

\item \textbf{Evidence-driven saturation failure in SMAP (\texttt{166\_SMAP\_id\_23}).}
This sequence is the key failure case in Stage~5's leave-one-out robustness in Appendix~\ref{sec:appendix-ablation-stages}. The level component and $s^{(1)}$ effectively detect the sharp level shift ($0.9753$ and $0.8214$), but latent novelty and channel pooling saturate after the transient. Their persistence beyond the $41$-sample anomaly lowers precision to $0.3049$, revealing a structural limitation of additive fusion: supplementary evidence can fail not by missing anomalies, but by remaining elevated after they have been detected.

\end{itemize}
  
\subsection{Computational Cost}
\label{sec:results-runtime}

Table~\ref{tab:runtime} reports the per-series wall-clock (fit plus scoring) and inference times for \LEAP{} and the five strongest competitors on each track, pooled over the held-out series and three random seeds ($540$ evaluations for TSB-AD-M and $1050$ for TSB-AD-U). Since runtime is heavy-tailed, mean $\pm$ standard deviation and the median (in parentheses) are reported. The divergence between these statistics is particularly pronounced for deep-learning methods, reflecting the disproportionate cost of high-dimensional and exceptionally long series.

\begin{table}[htbp]
\centering
\small
\setlength{\tabcolsep}{4pt}
\caption{Per-series wall-clock (fit plus scoring) and inference time in seconds, pooled over the held-out series and three seeds ($540$ runs on TSB-AD-M, $1050$ on TSB-AD-U). Mean $\pm$ std, with the median in parentheses; the mean training time is the difference between both columns. Dashes mark methods that are not among the five strongest competitors on that track.}
\label{tab:runtime}
\begin{tabular}{lcccc}
\toprule
 & \multicolumn{2}{c}{TSB-AD-M} & \multicolumn{2}{c}{TSB-AD-U} \\
\cmidrule(lr){2-3}\cmidrule(lr){4-5}
Method & Wall (s) & Infer (s) & Wall (s) & Infer (s) \\
\midrule
PaAno & $34.6 \pm 40.9$ (13.7) & $7.04 \pm 9.33$ (2.83) & $16.1 \pm 23.2$ (8.05) & $1.32 \pm 2.79$ (0.42) \\
PaAno$_{\mathrm{PAI}}$ & $14.5 \pm 8.05$ (13.1) & $1.53 \pm 3.03$ (0.89) & $13.0 \pm 6.76$ (10.4) & $0.56 \pm 1.19$ (0.20) \\
USAD & $24.6 \pm 27.6$ (13.6) & $7.41 \pm 7.90$ (3.81) & $9.74 \pm 14.7$ (3.45) & $2.52 \pm 5.40$ (0.83) \\
xLSTMAD-R & $3490 \pm 6112$ (966) & $154 \pm 169$ (66.3) & --- & --- \\
TimesNet & $55.4 \pm 68.6$ (25.6) & $11.4 \pm 14.3$ (5.23) & --- & --- \\
PatchTST & --- & --- & $21.1 \pm 31.6$ (7.65) & $5.68 \pm 12.1$ (1.89) \\
Sub-PCA & --- & --- & $3.84 \pm 19.8$ (0.21) & $0.00 \pm 0.00$ (0.00) \\
\midrule
\LEAP{} & $946 \pm 1007$ (566) & $95.8 \pm 192$ (31.2) & $93.4 \pm 82.6$ (75.8) & $17.2 \pm 28.8$ (9.11) \\
\bottomrule
\end{tabular}
\end{table}

Two patterns emerge. First, \LEAP{} incurs substantially higher total runtime than most lightweight baselines. On TSB-AD-M, its mean wall-clock time ($946$\,s) is $17$--$65$ times that of PaAno, PaAno$_{\mathrm{PAI}}$, USAD, and TimesNet; on TSB-AD-U ($93.4$\,s), it is $4$--$24$ times higher. The main exception is xLSTMAD-R, which is considerably slower than \LEAP{} ($3490$\,s vs.\ $946$\,s), although the gap narrows to $1.7$ times when comparing medians ($966$\,s vs.\ $566$\,s). Thus, \LEAP{} trades computational efficiency for improved detection performance, particularly on the multivariate track.

Second, training accounts for most of \LEAP{}'s runtime, with inference representing only $10.1\%$ of mean wall-clock time on TSB-AD-M and $18.4\%$ on TSB-AD-U. This profile is consistent with other learning-based methods, whereas non-learning approaches such as Sub-PCA incur negligible inference costs because most computation occurs during the initial matrix decomposition.


\section{Conclusions}
\label{sec:conclusions}

This paper introduced \LEAP{}, a time-series anomaly detector that replaces reconstruction with masked patch prediction, using prediction error to assess how well normal dynamics explain each observation, and augments it with three label-free evidence sources: density-calibrated latent novelty, local contrast, and per-channel novelty.

Under the TSB-AD protocol with VUS-PR as the primary metric, performance improves monotonically across the five stages, with every increment significant under a paired Wilcoxon signed-rank test. Since the evidence layers address largely disjoint subsets of series, their additive fusion preserves these gains; moreover, setting any fusion coefficient to zero reduces the model to its predecessor, providing a built-in controlled ablation. On the $180$ multivariate series, \LEAP{} achieves $0.5179 \pm 0.0022$, a $24.5\%$ improvement over the strongest locally measured baseline (PaAno), with posterior win probabilities above $99.9\%$ against the five strongest measured baselines. On the $350$ univariate series, the multi-scale variant combining masked-prediction scoring with latent novelty achieves $0.6194 \pm 0.0016$, outperforming all $20$ locally measured baselines, including PaAno$_{\mathrm{PAI}}$ ($+6.1\%$).

Two findings extend beyond the proposed architecture. First, the largest improvement came from evaluating prediction errors at the event rather than sample level. Second, the univariate advantage over PaAno$_{\mathrm{PAI}}$ is narrower, and PaAno$_{\mathrm{PAI}}$ retains stronger performance on AUC-PR, PA-F1, Event-based-F1 and Affiliation-F. Its performance benefits from benchmark-specific adaptations and post-hoc amplitude correction, highlighting the need to evaluate architectural complexity against strong baselines under robust metrics.

More broadly, the detectors that displaced reconstruction Transformers, such as PaAno, VACE and PAI-augmented models, converged on learning short subsequence representations and scoring their proximity to a bank of normal windows. \LEAP{} learns its representation through an objective that is itself a detector, so prediction error remains available as a second and independent signal, and the additive fusion exploits this redundancy.

Four directions follow from these results. A learned weighting of channels could replace the six fixed selection criteria of Eq.~\eqref{eq:criteria} and score each channel by its estimated reliability. Fusion coefficients that depend on observable properties of each series, such as its length, number of channels or dominant period, would move \LEAP{} toward the per-series adaptivity that benchmark heterogeneity calls for \cite{liu2024tsbad}. The uniform bank of $1{,}200$ windows could become a structured representation, such as the $k$-means coresets of Park\cite{park2026paano}. Finally, since the masked objective mirrors the pre-training of time-series foundation models \cite{goswami2024moment}, a pre-trained backbone combined with the same scoring stages would decouple representation from evidence aggregation and show whether the gains come from the representation or from the evidence fusion.

\section*{Acknowledgements}

This publication is part of the IAFER TSI-100927-2023-1 project, funded by the Recovery, Transformation, and Resilience Plan of the Next Generation of the EU through the Spanish Ministry of Digital Transformation and the Civil Service. This work was also supported by the Spanish Ministry of Science and Technology under grants PID2023-150070NB-I00 and PID2025-170986NA-I00 funded by MICIU/AEI/10.13039/501100011033 and by ``ERDF/EU'' and the ``European Union NextGenerationEU/PRTR'', respectively. This work was also supported by the University of Granada technology-transfer project OTRI-6915 (SmBuilding-X).

\section*{Data and code availability}
The datasets used in this study are publicly available as part of the TSB-AD
benchmark suite \cite{liu2024tsbad}. The implementation of \LEAP{} is
available at \url{https://github.com/ari-dasci/S-LEAP}.

\bibliographystyle{unsrt}  
\bibliography{references} 

\newpage
\appendix
\numberwithin{equation}{section}
\numberwithin{table}{section}
\numberwithin{figure}{section}

\section{Families of Time-Series Anomaly Detectors}
\label{app:background}

Table~\ref{tab:families} summarizes the families of time-series anomaly detectors reviewed in Section~\ref{sec:bg-tsad}, together with their scoring mechanism and the anomaly types to which they are suited.

\begin{table}[htbp]
\centering
\small
\caption{Families of time-series anomaly detectors, their scoring mechanism and
the anomaly types they are structurally suited to. \emph{Semi} denotes the
semi-supervised (clean training prefix) regime, \emph{Un} the unsupervised one.}
\label{tab:families}
\begin{tabular}{p{2.9cm}p{4.1cm}p{3.4cm}c}
\toprule
Family & Score mechanism & Suited to & Regime \\
\midrule
Statistical residual
  & residual of a fitted model (ARIMA, ETS, STL)
  & point, contextual & Semi/Un \\
Control charts, CPD
  & accumulated deviation of the mean
  & level shift, regime change & Semi/Un \\
Spectral
  & saliency of the amplitude spectrum
  & point, seasonal & Un \\
Discord/matrix profile
  & distance to nearest non-trivial match
  & subsequence, shape & Un \\
Normal-pattern banks
  & distance to a set of normal patterns
  & subsequence, shape & Semi \\
Forecasting (deep)
  & prediction residual
  & point, contextual & Semi \\
Reconstruction (deep)
  & reconstruction residual
  & point, subsequence & Semi/Un \\
Relational (graph)
  & deviation from predicted neighbor relations
  & relational & Semi \\
Contrastive SSL
  & discrepancy between views or neighborhoods
  & subsequence, contextual & Semi/Un \\
Representation $+$ bank
  & distance in a learned space to normal windows
  & subsequence, shape & Semi \\
Masked modeling
  & error of predicting removed content
  & shape, level, relational & Semi \\
\bottomrule
\end{tabular}
\end{table}

\section{Mathematical Notation Summary}
\label{app:notation}

Table~\ref{tab:notation} summarizes the notation used to define \LEAP{} in Section~\ref{sec:leap}.

\begin{table}[htbp]
\centering
\small
\caption{Notation used in Section~\ref{sec:leap}. Deployed values of all coefficients and widths are listed in Table~\ref{tab:hp}.}
\label{tab:notation}
\begin{tabular}{ll}
\toprule
Symbol & Meaning \\
\midrule
\multicolumn{2}{l}{\emph{Data, windowing and encoder}} \\
$T,\;C$ & series length, number of channels \\
$T_{\mathrm{tr}}$ & length of the anomaly-free training prefix \\
$x_t\in R^{C}$ & observation at time $t$ \\
$\tilde{x}_t$ & globally standardized observation \\
$\hat{P},\;\pi_{\mathrm{per}}$ & estimated dominant period, periods spanned by a window \\
$L,\;p,\;N=L/p$ & window length, patch length, patches per window \\
$D,\;H,\;B$ & model width, attention heads, encoder depth \\
$K$ & number of channel latents (bottleneck) \\
$h_{c,n}\in R^{D}$ & token of channel $c$, patch $n$ \\
$G,\;b$ & number of masking groups, mask block length (in patches) \\
\midrule
\multicolumn{2}{l}{\emph{Base scoring (Stage~1)}} \\
$e_{t,c}$ & masked prediction error at $(t,c)$ \\
$z_{t,c}$ & calibrated (standardized) error \\
$\lambda$ & shrinkage of the per-channel error spread toward the median of active channels \\
$\tau$ & temperature of the soft channel aggregation \\
$W_s=\eta\,p$ & event-scale smoothing width \\
$\kappa$ & tolerance-dilation width \\
$W_m,\;w_\ell$ & local-moment width, local-moment weight \\
\midrule
\multicolumn{2}{l}{\emph{Evidence layers (Stages~2--5)}} \\
$\mathcal{B},\;N_{\mathrm{bank}}$ & bank of training-window embeddings, bank size \\
$\chi$ & highest-variance channels in the concatenated embedding \\
$d(x)$ & latent novelty: distance of window $x$ to the bank \\
$\beta$ & latent-novelty weight (Stage~2) \\
$k,\;\alpha$ & density-calibration neighbors, local-density weight (Stage~3) \\
$\gamma,\;W_\gamma=\nu L$ & local-contrast weight, running-median width (Stage~4) \\
$\chi',\;\delta$ & channels per selection criterion, per-channel novelty weight (Stage~5) \\
\midrule
$unit(\bm{v})$ & $\bm{v}/q_{0.99}(\bm{v}\mid \bm{v}>0)$, per-series rescaling \\
$s^{(\cdot)}$ & score vector of a given stage \\
\bottomrule
\multicolumn{2}{l}{\footnotesize $p$ always denotes the patch length.
Statistical significance is}\\[-2pt]
\multicolumn{2}{l}{\footnotesize reported throughout as a $p$-value, never as a
bare $p$.}
\end{tabular}
\end{table}

\section{Algorithm, Complexity and Label-Freeness}
\label{app:algorithm}

Algorithm~\ref{alg:leap} summarizes the procedure.

\begin{algorithm}[htbp]
\caption{\LEAP{} fit and score}
\label{alg:leap}
\begin{algorithmic}[1]
\Statex \textbf{Input} series $X \in R^{T\times C}$, split $T_{\mathrm{tr}}$
\Statex \textbf{Output} score $s \in R^{T}$
\Statex \textit{Fit}
\State $(\mu,\tilde{\sigma},\mathcal{Q}) \gets$ standardizer on $X_{1:T_{\mathrm{tr}}}$ \Comment{Eqs.~\eqref{eq:anchor}--\eqref{eq:standardise}}
\State $L \gets$ window length from the dominant period \Comment{Eqs.~\eqref{eq:period}--\eqref{eq:winsize}}
\State train $f_\theta$ by masked patch prediction with early stopping \Comment{Eq.~\eqref{eq:loss}}
\State $(m_c, s^\star_c) \gets$ error calibration on training windows \Comment{Eqs.~\eqref{eq:calibmoments}--\eqref{eq:zscore}}
\State $(\mu^M,\sigma^M,\mu^S,\sigma^S) \gets$ local-moment calibration \Comment{Eq.~\eqref{eq:level}}
\State $\mathcal{B} \gets$ concatenated training bank with LOO distances \Comment{Eqs.~\eqref{eq:embed},~\eqref{eq:lof}}
\Statex \textit{Score}
\State $\bm{e} \gets$ masked error over the $G$ complementary passes \Comment{Eq.~\eqref{eq:maskederror}}
\State $\bm{z} \gets$ per-channel calibration of $\bm{e}$ \Comment{Eq.~\eqref{eq:zscore}}
\State $\bm{a} \gets$ soft-max channel aggregation of $\bm{z}$ \Comment{Eq.~\eqref{eq:chanagg}}
\State $s \gets smooth(\text{overlap-add}(\bm{a}))$ \Comment{Eqs.~\eqref{eq:overlapadd}--\eqref{eq:smooth}}
\State $\bm{\ell} \gets$ local-moment evidence \Comment{Eq.~\eqref{eq:level}}
\State $s^{(1)} \gets dil_\kappa\big(unit(s) + w_\ell unit(\bm{\ell})\big)$ \Comment{Eq.~\eqref{eq:stage1}}
\State $\hat{\tilde{\bm{d}}} \gets$ calibrated bank novelty with overlap-add \Comment{Eqs.~\eqref{eq:lof},~\eqref{eq:overlapadd}}
\State $s^{(3)} \gets unit(s^{(1)}) + \beta\,unit(\hat{\tilde{\bm{d}}})$ \Comment{Eqs.~\eqref{eq:lof},~\eqref{eq:stage3}}
\State $\mathcal{P} \gets$ union of the top-$\min(C,\chi')$ channels of six criteria \Comment{Eq.~\eqref{eq:criteria}}
\State $\mathcal{B}_c \gets$ training banks with LOO distances for $c\in\mathcal{P}$ \Comment{Eq.~\eqref{eq:lof}}
\State $\bm{A} \gets$ pooled per-channel novelty from $\mathcal{B}_c$ \Comment{Eq.~\eqref{eq:agg}}
\State $s^{(5a)} \gets s^{(3)} + \delta\,unit(\bm{A})$ \Comment{Eq.~\eqref{eq:stage5a}}
\State \Return $s^{(5a)} + \gamma\,unit_q(\text{contrast}(s^{(5a)}))$ \Comment{Eq.~\eqref{eq:stage5}}
\end{algorithmic}
\end{algorithm}

\paragraph{Complexity}

Let $W$ be the number of windows. A forward pass costs
$\mathcal{O}\!\left(B\left[(CN^2+NC^2)D+CND^2\right]\right)$ per window.
With the latent bottleneck, this becomes
$\mathcal{O}\!\left(B\left[(CN^2+NCK)D+CND^2\right]\right)$.
The $D^2$ term covers token projections and feed-forward layers.
Scoring uses $G=4$ passes per window. Nearest-neighbor queries use a
$k$-d tree over at most $1200$ bank vectors per channel considered. Their
cost depends on the embedding dimension and bank geometry and is at most
$\mathcal{O}(W(1+|\mathcal{P}|)|\mathcal{B}|)$ for brute-force search. The other
evidence layers add $\mathcal{O}(T)$ for local moments and
$\mathcal{O}(T\log W_\gamma)$ for the running median. In practice,
the Transformer passes dominate.

\paragraph{Label-Freeness}
Every quantity used by \LEAP{} depends only on $X$ and the known split
$T_{\mathrm{tr}}$. Standardization, window length, network weights, error
moments and bank embeddings use the training prefix. The channel criteria
of Eq.~\eqref{eq:criteria} compare it with the unlabeled remainder.
Unit rescaling uses the $99$th percentile of each score's own active
support. Ground-truth labels inform neither fitting nor per-series scoring. Global architecture and
hyperparameters were selected on the labeled tuning subset under the
protocol of Section~\ref{sec:exp-tuning}.

\section{TSB-AD Composition per Dataset}
\label{app:composition-per-dataset}

Table~\ref{tab:benchmarks} reports the number of held-out series and the application domains contributed by each source dataset of TSB-AD.

\begin{table}[htbp]
\centering
\small
\caption{Composition of the two TSB-AD held-out evaluation sets used in this
work (number of series per source dataset).}
\label{tab:benchmarks}
\begin{tabular}{lr@{\hskip 2em}lr}
\toprule
\multicolumn{2}{c}{\textbf{TSB-AD-M} (180 series)} &
\multicolumn{2}{c}{\textbf{TSB-AD-U} (350 series)} \\
\cmidrule(r){1-2}\cmidrule(l){3-4}
Source & \# & Source & \# \\
\midrule
SVDB        & 28 & UCR         & 70 \\
Exathlon    & 25 & SMD         & 33 \\
SMAP        & 25 & Exathlon    & 30 \\
GHL         & 23 & YAHOO       & 30 \\
SMD         & 20 & OPPORTUNITY & 27 \\
MSL         & 14 & NAB         & 23 \\
MITDB       & 11 & WSD         & 20 \\
TAO         & 11 & SVDB        & 18 \\
OPPORTUNITY &  7 & SMAP        & 17 \\
CATSv2      &  5 & IOPS        & 15 \\
LTDB        &  4 & TODS        & 13 \\
SWaT        &  2 & LTDB, MGAB, NEK, Stock & 8 each \\
CreditCard, Daphnet, GECCO, & & MITDB, MSL & 7 each \\
Genesis, PSM & 1 each & SED, TAO & 2 each \\
            &    & CATSv2, Daphnet, Power, SWaT & 1 each \\
\midrule
\multicolumn{2}{l}{\emph{Domains}: Sensor 71, Facility 46,} &
\multicolumn{2}{l}{\emph{Domains}: Facility 87, WebService 64,} \\
\multicolumn{2}{l}{Medical 43, Environment 11,} &
\multicolumn{2}{l}{Medical 47, HumanActivity 43, Sensor 39,} \\
\multicolumn{2}{l}{HumanActivity 8, Finance 1} &
\multicolumn{2}{l}{Synthetic 39, Environment 18, Finance 8, Traffic 5} \\
\bottomrule
\end{tabular}
\end{table}

\section{Hyperparameters}
\label{app:hyperparameters}

Table~\ref{tab:hp} summarizes the hyperparameters used for \LEAP{}.

\begin{table}[htbp]
\centering
\small
\caption{Hyperparameters of \LEAP{}. Adaptive entries are rules evaluated per
series. The track-specific rows are the only values that differ between the
multivariate and the univariate configuration; on the univariate track each
score is read out at every listed width and the maximum unit-rescaled response
is kept.}
\label{tab:hp}
\begin{tabular}{lllc}
\toprule
Group & Symbol & Meaning & Value \\
\midrule
\multirow{6}{*}{Architecture}
 & $p$ & patch length & $16$ \\
 & $D$ & model width & $128$ \\
 & $H$ & attention heads & $8$ \\
 & $B$ & encoder blocks & $4$ \\
 & $d_{\mathrm{ff}}$ & feed-forward width & $4D = 512$ \\
 & dropout & --- & $0.0$ \\
\midrule
\multirow{3}{*}{Channel axis}
 & $C_{\mathrm{lat}}$ & latent-bottleneck threshold & $128$ \\
 & $K$ & number of channel latents & $16$ \\
 & $\tau$ & channel-aggregation temperature & $0.5$ \\
\midrule
\multirow{2}{*}{Masking}
 & $G$ & masking groups (masked fraction $1/G$) & $4$ \\
 & $b$ & mask block, in patches & $1$ \\
\midrule
\multirow{4}{*}{Windowing (adaptive)}
 & $L$ & window length & $\min(2\hat{P}, 384)$, mult.\ of $p$ \\
 & $\Delta$ & window stride & $L/8$ \\
 & --- & min./max.\ training windows & $512$ / $4000$ \\
 & $\kappa$ & dilation width & $L/10$ \\
\midrule
\multirow{6}{*}{Optimization}
 & $\eta_0$ & learning rate (AdamW) & $5\times 10^{-4}$ \\
 & --- & weight decay & $10^{-4}$ \\
 & --- & batch size & $32$ \\
 & --- & max.\ epochs / patience & $400$ / $10$ \\
 & --- & LR decay factor / patience & $0.5$ / $3$ \\
 & --- & gradient clipping / validation split & $1.0$ / $0.2$ \\
\midrule
\multirow{5}{*}{Calibration}
 & $\rho_\sigma$ & degenerate-channel floor fraction & $0.05$ \\
 & $\lambda$ & error-spread shrinkage & $0.0$ \\
 & $\rho_f$ & error-spread floor fraction & $0.1$ \\
 & $\rho_\omega$ & edge-weight Gaussian fraction & $0.25$ \\
 & $\rho_\gamma$ & contrast denominator floor & $10^{-3}$ \\
\midrule
\multirow{7}{*}{Evidence fusion}
 & $w_\ell$ & local-moment weight (\texttt{add}) & $1.0$ \\
 & $\beta$ & latent-novelty weight & $1.0$ \\
 & $\chi$ & channels in the concat embedding & $4$ \\
 & $N_{\mathrm{bank}}$ & latent bank size & $1200$ \\
 & $k,\ \alpha$ & LOF neighbors, density weight & $4$, $0.25$ \\
 & $\nu$ & contrast width, $W_\gamma = \nu L$ & $64$ \\
 & $\chi'$ & channels per criterion in the pool & $4$ \\
\midrule
\multirow{4}{*}{Track-specific}
 & $W_s$ & score smoothing width(s) & $8p = 128$ (M) / $\{1, 8, 32, 128\}$ (U) \\
 & $W_m$ & local-moment width(s) & $\max(8, 4p) = 64$ (M) / $\{16, 64, 256\}$ (U) \\
 & $\gamma$ & contrast weight & $1.5$ (M) / $0$ (U) \\
 & $\delta$ & per-channel novelty weight & $1.5$ (M) / $0$ (U) \\
\bottomrule
\end{tabular}
\end{table}

\section{Hardware and Software Environment}
\label{sec:appendix-hardware}

All computational workloads were executed on an NVIDIA DGX-1 system equipped with dual Intel Xeon E5-2698 v4 processors, $512$\,GB of DDR4 RAM, and an NVIDIA Tesla V100 Tensor Core GPU ($32$\,GB VRAM). The software environment was configured on Ubuntu 22.04 LTS, Python 3.11, PyTorch v2.10.0, CUDA 12.8, and cuDNN v9.10.2.

All reported evaluation runs use standard FP32 precision. Under this configuration, a complete evaluation pass requires on average $545$\,minutes for TSB-AD-U ($350$ series) and $2{,}839$\,minutes for TSB-AD-M ($180$ series), consistent with Table~\ref{tab:runtime}, with peak GPU memory usage reaching $15$\,GB.

\section{Extended Empirical Ablation Analysis}
\label{sec:appendix-ablation}

This appendix presents an extended evaluation of the proposed architecture. We first report the stage-by-stage cumulative ablation across the full pipeline, followed by a micro-ablation isolating specific design decisions within the base detector.

\subsection{Stage-Wise Pipeline Ablation}
\label{sec:appendix-ablation-stages}

Table~\ref{tab:ablation-m} details performance progression across the five architectural stages on the 180 held-out multivariate series of TSB-AD-M. Because each stage strictly extends its predecessor, consecutive rows evaluate the isolated impact of a single mechanism under unified weighting, windowing, embedding, and calibration settings.

\begin{table}[htbp]
\centering
\small
\caption{Stage-wise ablation on TSB-AD-M (180 held-out series) single seed (2024). $\Delta$ denotes the mean per-series difference relative to the preceding stage; W/L/T indicates wins, losses, and ties; $p$-values correspond to two-sided Wilcoxon signed-rank tests on paired per-series VUS-PR scores.}
\label{tab:ablation-m}
\begin{tabular}{llccccc}
\toprule
Stage & Model & VUS-PR & $\Delta$ & W/L/T & $p$-value \\
\midrule
1 & Masked-patch base                       & 0.4611 & ---      & ---        & --- \\
2 & $+$ Latent novelty ($\beta{=}1.0$)      & 0.4954 & $+0.0343$ & 113/55/12 & $1.2\times10^{-5}$ \\
3 & $+$ Density calibration                 & 0.5034 & $+0.0080$ & 84/42/54  & $9.5\times10^{-5}$ \\
4 & $+$ Local contrast                      & 0.5074 & $+0.0040$ & 87/35/58  & $2.8\times10^{-6}$ \\
5 & $+$ Per-channel pool (\textbf{\LEAP{}}) & \textbf{0.5171} & $+0.0097$ & 90/51/39 & $3.4\times10^{-5}$ \\
\bottomrule
\end{tabular}
\end{table}

Four core insights emerge from this stage-wise breakdown:

\paragraph{Statistical Significance and Complementarity}
The overall gain from Stage~1 to Stage~5 reaches $+0.056$ VUS-PR (a $12\%$ relative improvement), composed of four distinct increments ranging between $+0.004$ and $+0.034$. This performance distribution underscores a modular design targeting largely disjoint failure modes: no single term universally dominates across all series, yet each addresses specific anomaly types and wins on more series than it loses. The high tie proportions in Stages~3 and 4 further reflect this behavior, as these mechanisms remain passive when local density or temporal context exhibits no anomaly signal. Importantly, gains overlap little at the individual series level: at a threshold of $\Delta \text{VUS-PR} \ge 0.01$, Stage~3 improves $35$ series and Stage~4 improves $43$ series, yet only $7$ series benefit from both by that margin, and only $3$ of the $180$ series improve across all four incremental stages. All four increments remain significant after Bonferroni correction for the four stage-wise comparisons (largest adjusted $p$-value $=3.8\times10^{-4}$).

\paragraph{Impact of Latent Novelty}
Stage~2 accounts for approximately $60\%$ of the total cumulative improvement. This outcome supports the theoretical motivation in Section~\ref{sec:leap-stage2}: local masked prediction can miss globally novel sequences that remain locally predictable. The deployed $\beta = 1.0$ was fixed a priori, assigning equal nominal weight to latent novelty and masked-prediction evidence, independently of held-out results. On the held-out set, $\beta = 0.6$ improved upon Stage~1 ($+0.0242$ VUS-PR, 111/54/15 W/L/T, $p$-value $= 5.7 \times 10^{-6}$), while increasing $\beta$ to $1.0$ yielded a further significant gain ($+0.0101$ VUS-PR, 104/56/20 W/L/T, $p$-value $= 1.7 \times 10^{-4}$). On the 20-series tuning set, $\beta = 0$ slightly outperformed the deployed $\beta = 1.0$ in mean VUS-PR (0.6761 vs. 0.6717). However, LOSO analysis reversed this preference, favoring $\beta = 1.0$ and revealing the sensitivity of the apparent $\beta = 0$ advantage to individual series. The deployed value was therefore consistent with both its a priori equal-weighting rationale and the tuning-set robustness analysis. Its substantial benefit was subsequently observed on the held-out set, which was not used for parameter selection.

\paragraph{Manifold Standardization via Density Calibration}
Stage~3 makes nearest-neighbor distances comparable across regions of varying density. Although $\alpha=1.0$ achieves the highest aggregate VUS-PR on the tuning set ($0.7101$, $\Delta=+0.0384$; Table~\ref{tab:sensitivity-coeffs}), its gains are concentrated in two series, with zero median improvement and degradation in $9$ of $20$ series. This instability near $\overline{\delta}_k \to 0$ motivates the conservative $\alpha=0.25$. 
Performance is also robust to neighborhood size: in a preliminary sweep on the $20$ tuning series, $k\in\{1,4,8\}$ improved over uncalibrated novelty by $+0.0144$, $+0.0123$, and $+0.0125$ VUS-PR, respectively, while on the held-out set the deployed $k=4$ contributes $+0.0080$ (Table~\ref{tab:ablation-m}).

\paragraph{Robustness Across Benchmark Series}
The performance improvement in Stage~5 ($+0.0097$ VUS-PR) is distributed broadly across the evaluation suite rather than driven by extreme outliers. A leave-one-series-out sensitivity analysis demonstrates that removing the most influential sequence (\texttt{166\_SMAP}) slightly increases the mean gain to $+0.0111$ VUS-PR, confirming the generalizability of per-channel pooling across the benchmark.

\subsection{Micro-Ablation of the Base Detector}
\label{sec:appendix-ablation-micro}

The Stage~1 baseline score ($s^{(1)}$ in Eq.~\eqref{eq:stage1}; VUS-PR $0.4611$) incorporates the score calibration, channel aggregation, event-scale smoothing, and local-moment formulations detailed in Section~\ref{sec:leap-scoring}, rather than raw reconstruction errors. Table~\ref{tab:micro} isolates the specific contributions of event-scale smoothing and local-moment evidence.

To disentangle individual design decisions, single-variable ablation experiments were conducted on the 20-series tuning set. Because each configuration is tested independently against its baseline, initial scores for smoothing ($0.4109$) and local moments ($0.4439$) differ. These development baselines also differ from the Stage~1 score ($0.4611$), which reflects final execution on the 180-series held-out evaluation set after locking hyperparameters during the tuning phase. The aggregation-rule comparison is additionally reported using Average Precision (AP), whereas the other micro-ablations use the primary VUS-PR metric. This component-level evaluation reveals two key findings:

\begin{table}[htbp]
\centering
\small
\caption{Single-variable micro-ablation of base detector design decisions on the tuning set.}
\label{tab:micro}
\begin{tabular}{llcc}
\toprule
Design Decision & Configuration Comparison & Metric & Performance \\
\midrule
Event-scale smoothing (Eq.~\eqref{eq:smooth})
  & $W_s = 1 \rightarrow W_s = 8p$
  & VUS-PR & $0.4109 \rightarrow 0.5135$ \\
Local-moment evidence (Eq.~\eqref{eq:level})
  & $w_\ell = 0 \rightarrow w_\ell > 0$
  & VUS-PR & $0.4439 \rightarrow 0.4539$ \\
Moment combination scheme
  & $\max$ vs.\ rescaled sum
  & AP & $0.3528 \rightarrow 0.3976$ \\
\bottomrule
\end{tabular}
\end{table}

\begin{enumerate}
    \item \textbf{Impact of Event-Scale Smoothing:} In alignment with Eq.~\eqref{eq:smooth}, event-scale smoothing serves as a primary performance driver in the base detector. Increasing the smoothing window from $W_s = 1$ to $W_s = 8p$ improves tuning VUS-PR from $0.4109$ to $0.5135$. This step establishes a solid scoring foundation upon which subsequent pipeline stages (Stages~2--5) achieve further cumulative gains across the held-out evaluation set ($+0.056$ VUS-PR total; see Table~\ref{tab:ablation-m}).
    
    \item \textbf{Contribution of Local-Moment Evidence:} Incorporating local-moment evidence (Section~\ref{sec:leap-moments}) yields a modest but statistically significant gain across the tuning set ($0.4439 \rightarrow 0.4539$ VUS-PR, $p$-value $= 0.021$). This confirms its theoretical function as a targeted mechanism for detecting sustained level or dispersion shifts while remaining passive elsewhere. Additionally, the rescaled-sum aggregation in Eq.~\eqref{eq:level} improves Average Precision (AP) over the standard maximum operator ($0.3528 \rightarrow 0.3976$), providing complementary evidence for the selected combination rule.
\end{enumerate}

\section{Impact of Data and Domain Characteristics}
\label{sec:appendix-domain-breakdown}

This section provides a granular breakdown of the per-series VUS-PR performance aggregated in Tables~\ref{tab:overall}--\ref{tab:ninemetric-u}. Specifically, we compare the three-seed measured scores of \LEAP{} against the five top-performing baselines identified for each track:
\begin{itemize}
    \item \textbf{Multivariate (M):} PaAno, PaAno$_{\mathrm{PAI}}$, USAD, xLSTMAD-R, and TimesNet.
    \item \textbf{Univariate (U):} PaAno, PaAno$_{\mathrm{PAI}}$, PatchTST, Sub-PCA, and USAD.
\end{itemize}
To ensure strict empirical consistency, this disaggregated evaluation is derived directly from the exact same experimental runs utilized for the main aggregate metrics.

\paragraph{Performance across Datasets and Application Domains}
Tables~\ref{tab:breakdown-m} and \ref{tab:breakdown-u} stratify the per-series scores of both tracks according to the $17$ and $23$ source datasets detailed in Table~\ref{tab:benchmarks}, as well as by their respective application domains. By examining this breakdown, we aim to investigate whether the central finding of Liu \cite{liu2024tsbad}---that no single anomaly detection method consistently dominates across diverse data sources---holds true when \LEAP{} is introduced to the benchmark.

\begin{table}[htbp]
\centering
\small
\caption{Mean VUS-PR by source dataset and by application domain on the
TSB-AD-M held-out set (180 series), for \LEAP{} and the five measured
baselines of Table~\ref{tab:ninemetric}. Values are the three-seed per-series
scores of Table~\ref{tab:overall}, pooled within each group; boldface marks
the best method per row.}
\label{tab:breakdown-m}
\begin{tabular}{lrcccccc}
\toprule
Source & $n$ & \LEAP{} & PaAno & PaAno$_{\mathrm{PAI}}$ & USAD & xLSTMAD-R & TimesNet \\
\midrule
SVDB & 28 & \textbf{0.6411} & 0.5544 & 0.4583 & 0.4489 & 0.3462 & 0.4669 \\
Exathlon & 25 & \textbf{0.9907} & 0.7773 & 0.8705 & 0.9659 & 0.9454 & 0.8449 \\
SMAP & 25 & \textbf{0.5980} & 0.5316 & 0.3713 & 0.2657 & 0.2377 & 0.2922 \\
GHL & 23 & \textbf{0.1830} & 0.0085 & 0.0114 & 0.0130 & 0.0336 & 0.0078 \\
SMD & 20 & 0.3295 & 0.3319 & 0.3218 & 0.3959 & 0.3877 & \textbf{0.4714} \\
MSL & 14 & \textbf{0.3715} & 0.2108 & 0.3103 & 0.3058 & 0.3427 & 0.2556 \\
MITDB & 11 & \textbf{0.4497} & 0.3823 & 0.2539 & 0.1327 & 0.1440 & 0.1803 \\
TAO & 11 & 0.7161 & 0.6654 & \textbf{0.8498} & 0.7369 & 0.7393 & 0.7662 \\
OPPORTUNITY & 7 & 0.1133 & 0.1457 & 0.1491 & \textbf{0.3941} & 0.3229 & 0.0679 \\
CATSv2 & 5 & \textbf{0.2498} & 0.0781 & 0.0816 & 0.1865 & 0.2211 & 0.1807 \\
LTDB & 4 & 0.5332 & \textbf{0.6161} & 0.5288 & 0.4903 & 0.4079 & 0.5121 \\
SWaT & 2 & \textbf{0.5219} & 0.2395 & 0.3162 & 0.4521 & 0.5072 & 0.1470 \\
CreditCard & 1 & \textbf{0.0663} & 0.0256 & 0.0496 & 0.0636 & \textbf{0.0663} & 0.0585 \\
Daphnet & 1 & \textbf{0.3842} & 0.2395 & 0.3351 & 0.1260 & 0.2361 & 0.3393 \\
GECCO & 1 & 0.1030 & 0.1934 & 0.1007 & 0.1233 & 0.1923 & \textbf{0.3129} \\
Genesis & 1 & 0.7286 & 0.3141 & \textbf{0.8998} & 0.0488 & 0.6535 & 0.0170 \\
PSM & 1 & 0.2344 & 0.2070 & 0.1662 & 0.1685 & \textbf{0.3868} & 0.1798 \\
\midrule
\multicolumn{8}{l}{\emph{By application domain}} \\
\midrule
Sensor & 71 & \textbf{0.3871} & 0.2509 & 0.2244 & 0.1864 & 0.2039 & 0.1773 \\
Facility & 46 & 0.6868 & 0.5713 & 0.6166 & \textbf{0.7007} & 0.6908 & 0.6680 \\
Medical & 43 & \textbf{0.5821} & 0.5161 & 0.4126 & 0.3718 & 0.3002 & 0.3978 \\
Environment & 11 & 0.7161 & 0.6654 & \textbf{0.8498} & 0.7369 & 0.7393 & 0.7662 \\
HumanActivity & 8 & 0.1471 & 0.1574 & 0.1724 & \textbf{0.3606} & 0.3120 & 0.1018 \\
Finance & 1 & \textbf{0.0663} & 0.0256 & 0.0496 & 0.0636 & \textbf{0.0663} & 0.0585 \\
\bottomrule
\end{tabular}
\end{table}

On the TSB-AD-M track, \LEAP{} achieves the best performance on nine of the $17$ source datasets and ties with xLSTMAD-R on CreditCard. The remaining seven datasets are led by five competitors: TimesNet on GECCO and SMD, PaAno$_{\mathrm{PAI}}$ on Genesis and TAO, PaAno on LTDB, USAD on OPPORTUNITY, and xLSTMAD-R on PSM. This distribution mirrors the dataset fragmentation documented by Liu \cite{liu2024tsbad}, confirming that performance heterogeneity persists even against a top-performing detector.

Two datasets exhibit notable performance shifts. On GHL, \LEAP{} achieves its largest margin of superiority, outperforming the second-best model, xLSTMAD-R, by $+0.1494$. Where all competing methods fail to exceed a score of $0.034$, \LEAP{} reaches $0.183$. This suggests that integrating local-moment and latent-novelty components (Section~\ref{sec:leap-moments}) extracts actionable anomaly signals from regimes where standard detectors fail completely.

Conversely, the OPPORTUNITY dataset accounts for \LEAP{}'s largest relative deficit, falling $-0.2808$ behind USAD. Because OPPORTUNITY comprises $7$ of the $8$ series in the HumanActivity domain within the held-out set, this single dataset dictates the domain aggregate. As a result, \LEAP{} ranks fifth out of six methods in HumanActivity, despite leading two of the remaining five domains (Medical and Sensor, with Sensor being the largest domain at $71$ series).

Domain-level aggregation does not smooth out this underlying per-source fragmentation. \LEAP{} leads outright in two of the six domains and ties for first in a third (Finance, comprising a single series), whereas USAD dominates Facility and HumanActivity, and PaAno$_{\mathrm{PAI}}$ leads Environment. This breakdown confirms that the ``no free lunch'' trade-off observed at the dataset level persists after domain grouping.

\begin{table}[htbp]
\centering
\small
\caption{Mean VUS-PR by source dataset and by application domain on the
TSB-AD-U held-out set (350 series), for \LEAP{} and the five measured
baselines of Table~\ref{tab:ninemetric-u}. Values are the three-seed
per-series scores of Table~\ref{tab:overall-u}, pooled within each group;
boldface marks the best method per row.}
\label{tab:breakdown-u}
\begin{tabular}{lrcccccc}
\toprule
Source & $n$ & \LEAP{} & PaAno & PaAno$_{\mathrm{PAI}}$ & PatchTST & Sub-PCA & USAD \\
\midrule
UCR & 70 & \textbf{0.4593} & 0.3739 & 0.4017 & 0.2520 & 0.1237 & 0.1045 \\
SMD & 33 & \textbf{0.8270} & 0.5005 & 0.5327 & 0.8065 & 0.4416 & 0.5439 \\
Exathlon & 30 & \textbf{0.9626} & 0.8186 & 0.9285 & 0.8559 & 0.8920 & 0.9364 \\
YAHOO & 30 & 0.4170 & 0.6810 & \textbf{0.6968} & 0.2731 & 0.1547 & 0.1027 \\
OPPORTUNITY & 27 & 0.6304 & 0.2646 & 0.7673 & 0.0726 & \textbf{0.9139} & 0.7977 \\
NAB & 23 & \textbf{0.6446} & 0.4982 & 0.4355 & 0.4868 & 0.3947 & 0.4946 \\
WSD & 20 & \textbf{0.6690} & 0.4275 & 0.5153 & 0.3032 & 0.1229 & 0.0323 \\
SVDB & 18 & 0.6386 & \textbf{0.7045} & 0.5074 & 0.6158 & 0.5222 & 0.4422 \\
SMAP & 17 & 0.8043 & 0.8045 & \textbf{0.8566} & 0.7795 & 0.5045 & 0.4631 \\
IOPS & 15 & \textbf{0.5254} & 0.2673 & 0.4308 & 0.3103 & 0.2478 & 0.1531 \\
TODS & 13 & 0.6242 & 0.7296 & \textbf{0.7789} & 0.7160 & 0.5394 & 0.5493 \\
LTDB & 8 & 0.6172 & \textbf{0.6950} & 0.6326 & 0.5514 & 0.5558 & 0.5504 \\
MGAB & 8 & \textbf{0.3085} & 0.2831 & 0.1873 & 0.0677 & 0.0052 & 0.0056 \\
NEK & 8 & 0.6929 & 0.5662 & 0.7603 & 0.6322 & \textbf{0.8856} & 0.7237 \\
Stock & 8 & 0.6351 & 0.5538 & 0.6790 & 0.6274 & \textbf{0.7631} & 0.6197 \\
MITDB & 7 & 0.3953 & \textbf{0.4296} & 0.3200 & 0.2579 & 0.3101 & 0.2174 \\
MSL & 7 & 0.5639 & 0.2515 & 0.4915 & \textbf{0.6146} & 0.5085 & 0.4832 \\
SED & 2 & 0.7558 & \textbf{0.9531} & 0.9319 & 0.0891 & 0.0314 & 0.0351 \\
TAO & 2 & 0.9314 & 0.8704 & \textbf{0.9493} & 0.9308 & 0.9299 & 0.9283 \\
CATSv2 & 1 & 0.2872 & 0.3209 & \textbf{0.3743} & 0.3692 & 0.2647 & 0.3186 \\
Daphnet & 1 & 0.5049 & 0.3034 & 0.3543 & \textbf{0.5573} & 0.4218 & 0.4457 \\
Power & 1 & 0.1169 & \textbf{0.1857} & 0.1513 & 0.0718 & 0.0761 & 0.1650 \\
SWaT & 1 & \textbf{0.5346} & 0.0961 & 0.1786 & 0.0991 & 0.3930 & 0.3546 \\
\midrule
\multicolumn{8}{l}{\emph{By application domain}} \\
\midrule
Facility & 87 & \textbf{0.7499} & 0.5637 & 0.5953 & 0.6926 & 0.5359 & 0.5968 \\
WebService & 64 & \textbf{0.5500} & 0.4511 & 0.5379 & 0.3279 & 0.2976 & 0.2193 \\
Medical & 47 & 0.6345 & \textbf{0.6483} & 0.5704 & 0.4430 & 0.3956 & 0.3177 \\
HumanActivity & 43 & 0.5514 & 0.2445 & 0.5675 & 0.0989 & \textbf{0.6092} & 0.5365 \\
Sensor & 39 & \textbf{0.6998} & 0.6407 & 0.6967 & 0.5732 & 0.4073 & 0.3968 \\
Synthetic & 39 & 0.5391 & \textbf{0.6341} & 0.6324 & 0.4367 & 0.2423 & 0.2414 \\
Environment & 18 & 0.3288 & 0.2718 & \textbf{0.4108} & 0.3907 & 0.1370 & 0.1262 \\
Finance & 8 & 0.6351 & 0.5538 & 0.6790 & 0.6274 & \textbf{0.7631} & 0.6197 \\
Traffic & 5 & \textbf{0.6984} & 0.5691 & 0.4542 & 0.4339 & 0.3339 & 0.4983 \\
\bottomrule
\end{tabular}
\end{table}

On the TSB-AD-U track, performance fragmentation is even more pronounced. \LEAP{} leads on $8$ of the $23$ source datasets (UCR, SMD, Exathlon, NAB, WSD, IOPS, MGAB, and SWaT), followed by PaAno$_{\mathrm{PAI}}$ and PaAno on $5$ each, Sub-PCA on $3$, and PatchTST on $2$ (with USAD leading none). Consequently, even the top-ranked method secures the best score on barely more than a third of the available sources.

\LEAP{}'s two largest univariate deficits occur on OPPORTUNITY ($0.6304$ compared to $0.9139$ for Sub-PCA, $\Delta=-0.2835$) and YAHOO ($0.4170$ compared to $0.6968$ for PaAno$_{\mathrm{PAI}}$, $\Delta=-0.2798$). The YAHOO gap directly reflects the per-series instability previously noted in Figure~\ref{fig:overall-boxplot-u}. Specifically, $26$ of the $30$ YAHOO series are dominated by point-scale anomalies (Appendix~\ref{sec:results-type})---the exact anomaly category where \LEAP{} demonstrates the lowest relative efficacy, as detailed in the subsequent type-level breakdown.

The OPPORTUNITY deficit, however, cannot be attributed to the same morphological weakness observed in YAHOO. In OPPORTUNITY, the series are dominated by level shifts ($18$ of $27$) and dispersion changes ($7$ of $27$), two categories on which \LEAP{} leads every baseline on average (Table~\ref{tab:breakdown-type-u}). This discrepancy highlights a dataset-specific effect that type-level stratification cannot fully explain, serving as a reminder that categorical breakdowns capture average tendencies rather than guarantees for individual data sources.

Domain-level aggregation reflects a similarly mixed performance profile. \LEAP{} leads four of the nine domains, including the two largest, Facility ($87$ series) and WebService ($64$ series), together with Sensor and Traffic. In WebService, the YAHOO deficit is outweighed by the leads on WSD and IOPS ($+0.0121$ over PaAno$_{\mathrm{PAI}}$). Conversely, it trails Sub-PCA by $-0.1280$ in Finance and by $-0.0578$ in HumanActivity, PaAno by $-0.0950$ in Synthetic and narrowly by $-0.0138$ in Medical, and PaAno$_{\mathrm{PAI}}$ by $-0.0820$ in Environment. From a methodological standpoint, it is important to note that datasets such as NAB, UCR, and YAHOO contribute series to multiple domains under the standard TSB-AD taxonomy (with NAB alone spanning five). Consequently, the source-dataset and domain-level views represent genuinely distinct partitions of the $350$ series, rather than simple hierarchical coarsenings of one another.

\paragraph{By Anomaly Type}
\label{sec:results-type}

To evaluate model performance across fine-grained anomaly behaviors, time series must be stratified by structural anomaly type. While TSB-AD provides collection-level tags distinguishing sequence-only (\texttt{Seq}) from mixed point and sequence (\texttt{P\&Seq}) datasets \cite{liu2024tsbad}, this metadata is insufficient for granular ablation. Specifically, the official tags are assigned per collection rather than per individual time series, and they collapse distinct structural phenomena (including level shifts, dispersion changes, shape anomalies, and relational disruptions) into a single generic sequence category.

To establish a precise and reproducible classification, we operationalize the six-way taxonomy from Section~\ref{sec:bg-tsad} through an automated feature-extraction procedure applied directly to ground-truth anomaly segmentations. For each labeled anomalous segment, the target channel exhibiting the largest standardized mean shift relative to its preceding normal history is identified. Short segments containing five or fewer samples are classified as point-scale anomalies, categorized as \emph{global} ($z \geq 4$ relative to training marginals) or \emph{contextual} ($z \geq 3$ relative to local pre-event windows).

Longer anomalous segments are evaluated across three descriptive statistics computed on the primary channel: standardized mean shift (\emph{level}), log-ratio of standard deviations (\emph{dispersion}), and z-normalized Euclidean distance to nearest-matching normal reference windows (\emph{shape}). Segments are assigned to whichever category exceeds its calibrated threshold ($3.0$, $2$ octaves, and $2.5$, respectively) by the largest margin. For multivariate series, a segment is classified as \emph{relational} when every individual channel among the top $20$ highest-variance channels remains within normal bounds ($z < 2.5$) while the joint Mahalanobis distance exceeds $3.0$, isolating instances where inter-channel correlations break despite normal univariate marginals.

Finally, a time series is designated with the anomaly type that accounts for at least $60\%$ of its total anomalous duration, or classified as \emph{mixed} if no single category dominates. This automated classification establishes the per-series taxonomy utilized throughout Tables~\ref{tab:breakdown-type-m}--\ref{tab:breakdown-type-u}.

\begin{table}[htbp]
\centering
\small
\caption{Mean VUS-PR by automatically-assigned dominant anomaly type
(Appendix~\ref{sec:results-type}) on the TSB-AD-M held-out set, for \LEAP{}
and the five measured baselines of Table~\ref{tab:ninemetric}. Boldface marks
the best method per row.}
\label{tab:breakdown-type-m}
\begin{tabular}{lrcccccc}
\toprule
Type & $n$ & \LEAP{} & PaAno & PaAno$_{\mathrm{PAI}}$ & USAD & xLSTMAD-R & TimesNet \\
\midrule
Level shift & 88 & \textbf{0.5662} & 0.4455 & 0.4531 & 0.5097 & 0.4875 & 0.4466 \\
Subsequence/shape & 34 & \textbf{0.5265} & 0.4890 & 0.3730 & 0.3197 & 0.2574 & 0.3563 \\
Relational & 16 & \textbf{0.3547} & 0.1781 & 0.1707 & 0.1419 & 0.2146 & 0.1476 \\
Dispersion change & 16 & \textbf{0.3426} & 0.2675 & 0.2225 & 0.1561 & 0.1619 & 0.1593 \\
Mixed & 14 & \textbf{0.4573} & 0.3277 & 0.3261 & 0.2746 & 0.2845 & 0.3512 \\
Point (global) & 8 & 0.5828 & 0.5217 & \textbf{0.7148} & 0.6043 & 0.6079 & 0.6315 \\
Point (contextual) & 4 & 0.8202 & 0.7929 & \textbf{0.9198} & 0.8338 & 0.8336 & 0.8589 \\
\bottomrule
\end{tabular}
\end{table}

\begin{table}[htbp]
\centering
\small
\caption{Mean VUS-PR by automatically-assigned dominant anomaly type
(Appendix~\ref{sec:results-type}) on the TSB-AD-U held-out set, for \LEAP{}
and the five measured baselines of Table~\ref{tab:ninemetric-u}. Relational
anomalies do not apply to a univariate series and are absent by
construction. Boldface marks the best method per row.}
\label{tab:breakdown-type-u}
\begin{tabular}{lrcccccc}
\toprule
Type & $n$ & \LEAP{} & PaAno & PaAno$_{\mathrm{PAI}}$ & PatchTST & Sub-PCA & USAD \\
\midrule
Level shift & 115 & \textbf{0.8072} & 0.5183 & 0.7578 & 0.5954 & 0.7053 & 0.6768 \\
Subsequence/shape & 101 & \textbf{0.5255} & 0.4934 & 0.4102 & 0.4250 & 0.2243 & 0.2311 \\
Dispersion change & 68 & \textbf{0.5494} & 0.5102 & 0.5108 & 0.3323 & 0.3114 & 0.3109 \\
Mixed & 22 & \textbf{0.5895} & 0.4862 & 0.5796 & 0.4912 & 0.4823 & 0.4132 \\
Point (global) & 29 & 0.4455 & 0.6773 & \textbf{0.7188} & 0.3593 & 0.2403 & 0.1619 \\
Point (contextual) & 15 & \textbf{0.5083} & 0.4213 & 0.4965 & 0.3140 & 0.2347 & 0.1652 \\
\bottomrule
\end{tabular}
\end{table}

Categorizing performance by anomaly type evaluates \LEAP{}'s scoring components directly against their targeted failure modes rather than aggregate metrics. Across both benchmark tracks, \LEAP{} leads on level shifts, dispersion changes, shape anomalies, and mixed series, confirming the specific contributions of the local-moment term (Section~\ref{sec:leap-moments}), local-density calibration (Section~\ref{sec:leap-stage3}), and latent discord scoring (Section~\ref{sec:leap-stage2}). Furthermore, on TSB-AD-M, \LEAP{} achieves its widest margin of advantage on relational anomalies ($+0.1401$ VUS-PR over xLSTMAD-R), demonstrating that group-masking (Section~\ref{sec:leap-masking}) effectively exposes broken cross-channel dependencies.

Conversely, \LEAP{} displays a clear sensitivity limitation on point-scale anomalies. On TSB-AD-M, it ranks near the bottom of the field on both global and contextual point events, reaching $0.5828$ and $0.8202$ compared to PaAno$_{\mathrm{PAI}}$'s $0.7148$ and $0.9198$. On TSB-AD-U, \LEAP{} leads narrowly on contextual point anomalies ($0.5083$ versus $0.4965$ for PaAno$_{\mathrm{PAI}}$), but global point anomalies remain its weakest category, ranking third of six ($0.4455$ versus $0.7188$ for PaAno$_{\mathrm{PAI}}$ and $0.6773$ for PaAno). Structurally, masked-patch prediction, moving averages, and $k$-nearest-neighbor novelty metrics operate over patch-wide windows, rendering them inherently less sensitive to isolated sample spikes than density (or memory-bank) based scorers. Because point-scale anomalies represent a small subset of the evaluation sets ($12$ of $180$ series on M; $44$ of $350$ on U), this failure mode has a constrained impact on aggregate scores, yet it establishes the primary operational limit of \LEAP{}'s design.

\paragraph{By Series Characteristics}

To evaluate the empirical necessity and behavior of the adaptive hyper-parameter rules (Table~\ref{tab:hp}), per-series evaluation scores are stratified across four intrinsic properties: channel count (multivariate track only), sequence length, anomaly ratio, and the presence of a dominant period estimated via Eqs.~\eqref{eq:period}--\eqref{eq:acfband} on the training prefix. Tables~\ref{tab:breakdown-char-m} and \ref{tab:breakdown-char-u} summarize these performance breakdowns on TSB-AD-M and TSB-AD-U.

\begin{table}[htbp]
\centering
\small
\caption{Mean VUS-PR by series characteristics on the TSB-AD-M held-out set,
for \LEAP{} and the five measured baselines of Table~\ref{tab:ninemetric}; boldface
marks the best method per row.}
\label{tab:breakdown-char-m}
\begin{tabular}{lrcccccc}
\toprule
& $n$ & \LEAP{} & PaAno & PaAno$_{\mathrm{PAI}}$ & USAD & xLSTMAD-R & TimesNet \\
\midrule
\multicolumn{8}{l}{\emph{By number of channels}} \\
\midrule
2--10 & 58 & \textbf{0.6089} & 0.5505 & 0.5089 & 0.4524 & 0.4028 & 0.4587 \\
11--30 & 72 & \textbf{0.5182} & 0.3660 & 0.3356 & 0.3291 & 0.3333 & 0.3185 \\
31--248 & 50 & 0.4121 & 0.3322 & 0.3827 & \textbf{0.4514} & 0.4499 & 0.4037 \\
\midrule
\multicolumn{8}{l}{\emph{By series length (samples)}} \\
\midrule
$\leq10{,}000$ & 53 & \textbf{0.5424} & 0.4601 & 0.4517 & 0.3867 & 0.3812 & 0.3720 \\
$10{,}001$--$46{,}655$ & 38 & 0.5043 & 0.4415 & 0.4719 & 0.5296 & \textbf{0.5391} & 0.5059 \\
$46{,}656$--$200{,}000$ & 55 & \textbf{0.4932} & 0.3334 & 0.3293 & 0.3532 & 0.3443 & 0.3308 \\
$>200{,}000$ & 34 & \textbf{0.5351} & 0.4527 & 0.3772 & 0.3666 & 0.3010 & 0.3703 \\
\midrule
\multicolumn{8}{l}{\emph{By anomaly ratio}} \\
\midrule
$\leq0.0114$ & 45 & \textbf{0.4094} & 0.2795 & 0.2427 & 0.1857 & 0.1993 & 0.2091 \\
$0.0115$--$0.0374$ & 45 & \textbf{0.3830} & 0.3038 & 0.2221 & 0.2243 & 0.2020 & 0.2422 \\
$0.0375$--$0.0715$ & 45 & \textbf{0.5584} & 0.4626 & 0.5092 & 0.5259 & 0.4928 & 0.4876 \\
$>0.0715$ & 45 & \textbf{0.7210} & 0.6184 & 0.6440 & 0.6754 & 0.6583 & 0.6105 \\
\midrule
\multicolumn{8}{l}{\emph{By dominant period (Eq.~\eqref{eq:period})}} \\
\midrule
Period found & 176 & \textbf{0.5109} & 0.4072 & 0.3923 & 0.3930 & 0.3779 & 0.3767 \\
No period found & 4 & 0.8293 & 0.8062 & \textbf{0.9408} & 0.8352 & 0.8392 & 0.8556 \\
\bottomrule
\end{tabular}
\end{table}

\begin{table}[htbp]
\centering
\small
\caption{Mean VUS-PR by series characteristics on the TSB-AD-U held-out set,
for \LEAP{} and the five measured baselines of Table~\ref{tab:ninemetric-u}.
Every series is univariate by construction, so the channel stratification of
Table~\ref{tab:breakdown-char-m} does not apply. Boldface marks the best method per
row.}
\label{tab:breakdown-char-u}
\begin{tabular}{lrcccccc}
\toprule
& $n$ & \LEAP{} & PaAno & PaAno$_{\mathrm{PAI}}$ & PatchTST & Sub-PCA & USAD \\
\midrule
\multicolumn{8}{l}{\emph{By series length (samples)}} \\
\midrule
$\leq7{,}500$ & 88 & 0.5785 & 0.5866 & \textbf{0.6334} & 0.4839 & 0.3915 & 0.3733 \\
$7{,}501$--$18{,}228$ & 87 & 0.6215 & 0.4511 & \textbf{0.6430} & 0.3175 & 0.3946 & 0.3296 \\
$18{,}229$--$43{,}066$ & 96 & \textbf{0.7434} & 0.5540 & 0.6254 & 0.6177 & 0.5124 & 0.5459 \\
$>43{,}066$ & 79 & \textbf{0.5119} & 0.4651 & 0.4130 & 0.3849 & 0.3553 & 0.3118 \\
\midrule
\multicolumn{8}{l}{\emph{By anomaly ratio}} \\
\midrule
$\leq0.0048$ & 88 & \textbf{0.4496} & 0.4402 & 0.4205 & 0.2330 & 0.0881 & 0.0437 \\
$0.0049$--$0.0215$ & 87 & \textbf{0.6515} & 0.5124 & 0.5845 & 0.5153 & 0.2979 & 0.3023 \\
$0.0216$--$0.0774$ & 87 & 0.6349 & 0.5053 & \textbf{0.6427} & 0.4904 & 0.6166 & 0.5575 \\
$>0.0774$ & 88 & \textbf{0.7421} & 0.6080 & 0.6885 & 0.5900 & 0.6674 & 0.6810 \\
\midrule
\multicolumn{8}{l}{\emph{By dominant period (Eq.~\eqref{eq:period})}} \\
\midrule
Period found & 326 & \textbf{0.6106} & 0.5132 & 0.5714 & 0.4432 & 0.4059 & 0.3893 \\
No period found & 24 & 0.7388 & 0.5620 & \textbf{0.7532} & 0.6426 & 0.5709 & 0.4855 \\
\bottomrule
\end{tabular}
\end{table}

Three primary empirical trends emerge from this stratification. First, \LEAP{}'s advantage on TSB-AD-M does not extend to the widest series. It leads in the $2$--$10$ channel band ($+0.0584$ VUS-PR over PaAno) and most clearly in the $11$--$30$ band ($+0.1522$ over PaAno), but falls behind USAD and xLSTMAD-R in the $31$--$248$ band ($-0.0393$ relative to USAD). Because the per-channel pooling step (Section~\ref{sec:leap-stage5}) selects a representative channel subset rather than modeling all channels jointly, its selectivity degrades as the candidate channel pool expands. This empirical drop directly supports replacing fixed pooling criteria with learned reliability weighting (Section~\ref{sec:conclusions}).

Second, sequence length has a minimal impact on overall rankings. On the multivariate track, \LEAP{} ranks first in three of the four length bins and fourth in the $10{,}001$--$46{,}655$ bin, where xLSTMAD-R, USAD and TimesNet lead by at most $0.035$. On the univariate track, it leads the two longest bins outright, ranks second in the $7{,}501$--$18{,}228$ bin, and drops to third of six in the shortest bin ($0.5785$ versus $0.6334$ for PaAno$_{\mathrm{PAI}}$, $\Delta=-0.0549$). Unlike channel count, this degradation is modest and confined to the short-series half of the set, where PaAno$_{\mathrm{PAI}}$ leads, establishing sequence length as a secondary regime driver.

Third, absolute scores across all evaluated methods scale positively with the anomaly ratio, as ranking-based metrics inherently benefit from higher positive sample density (only $5$ minor dips under $0.03$ occur across $36$ transitions between consecutive anomaly-ratio bins). However, \LEAP{}'s advantage varies across tracks. On TSB-AD-M, its largest margin over the runner-up occurs in the lowest anomaly-ratio bin ($+0.1299$), decreasing to $+0.0792$, $+0.0325$, and $+0.0456$ in the remaining bins, highlighting its strength on sparse anomalies. On TSB-AD-U, \LEAP{} also leads the sparsest bin, but only narrowly ($+0.0094$ over PaAno); its largest margin falls in the second bin ($+0.0670$ over PaAno$_{\mathrm{PAI}}$), it trails PaAno$_{\mathrm{PAI}}$ marginally in the third ($-0.0078$), and it leads again in the densest band ($+0.0536$). The narrow univariate margin at low density aligns with structural differences in anomaly types across tracks. Sparse univariate anomalies are disproportionately point-scale (mean ratio of $0.017$ versus $0.049$ for other types), exposing \LEAP{}'s point-level sensitivity limitations. Conversely, multivariate point anomalies exhibit higher average density ($0.079$ versus $0.048$ for other types), and the lowest multivariate ratio bin consists predominantly ($80\%$) of level-shift, dispersion, or relational events, which directly match \LEAP{}'s additive evidence criteria.

The dominant-period split requires an important caveat. All four multivariate series lacking a detected period belong to TAO recordings, whose anomaly ratios ($0.094$--$0.141$) fall in the highest anomaly-ratio bin ($>0.0715$). Similarly, the $24$ univariate series without a period display a higher mean anomaly ratio ($0.061$ versus $0.043$). Because every method scores substantially higher on the no-period subset, this performance gap is confounded by anomaly density rather than reflecting an inherent advantage of the fallback window rule ($L=\min(L_{\max},16p)$ in Eq.~\eqref{eq:winsize}). These results confirm that the fallback rule introduces no empirical liability, though the small sample sizes ($4$ and $24$ series, respectively) preclude drawing stronger conclusions in either direction.

\section{Extended Sensitivity Analysis}
\label{app:sensitivity}

This appendix complements Section~\ref{sec:results-sensitivity} by evaluating the two pipeline combination rules and the five architectural backbone constants.

\subsection{Combination Rules}
\label{app:sensitivity-rules}

The deployed architecture fixes two pipeline combination rules: local-moment fusion (\texttt{level\_combine}) and channel aggregation of the calibrated prediction errors (\texttt{chan\_agg}). Both were evaluated against candidate alternatives on the tuning set (Table~\ref{tab:combine-rules}).

\begin{table}[htbp]
\centering
\small
\caption{Effect of alternative combination rules for local-moment fusion (\texttt{level\_combine}) and channel aggregation (\texttt{chan\_agg}) on the $20$-series tuning set. $\Delta$ denotes the mean per-series VUS-PR difference relative to the deployed rule; Range indicates the minimum and maximum per-series $\Delta$; $p$-value is from a paired Wilcoxon signed-rank test; W/L/T reports wins, losses, and ties relative to the deployed baseline at a $0.001$ VUS-PR threshold.}
\label{tab:combine-rules}
\begin{tabular}{llccccc}
\toprule
Component & Rule & VUS-PR & $\Delta$ & Range of $\Delta$ & $p$-value & W/L/T \\
\midrule
\multirow{4}{*}{\texttt{level\_combine}}
 & \texttt{add} (deployed) & $0.6717$ & n/a & n/a & n/a & n/a \\
 & \texttt{pmean} & $0.6715$ & $-0.0002$ & $[-0.022, +0.019]$ & $0.275$ & $4/7/9$ \\
 & \texttt{max}   & $0.6725$ & $+0.0008$ & $[-0.041, +0.091]$ & $0.395$ & $5/10/5$ \\
 & \texttt{gate}  & $0.6571$ & $-0.0146$ & $[-0.155, +0.017]$ & $0.374$ & $8/8/4$ \\
\midrule
\multirow{3}{*}{\texttt{chan\_agg}}
 & \texttt{softmax} (deployed) & $0.6717$ & n/a & n/a & n/a & n/a \\
 & \texttt{max}  & $0.6717$ & $+0.0000$ & $[-0.0001, +0.0004]$ & $0.135$ & $0/0/20$ \\
 & \texttt{mean} & $0.6724$ & $+0.0007$ & $[-0.017, +0.013]$ & $0.154$ & $6/3/11$ \\
\bottomrule
\end{tabular}
\end{table}

Three of the four \texttt{level\_combine} rules achieve similar overall means, while \texttt{gate} trails by 0.0146. However, their per-series stability varies substantially. The \texttt{max} operator exhibits considerable variability across individual series ($-0.041$ to $+0.091$ VUS-PR) and achieves a marginal mean gain ($+0.0008$), despite recording twice as many losses as wins ($5/10/5$). The \texttt{gate} mechanism ranges from $-0.155$ to $+0.017$ VUS-PR, producing the largest mean decline ($-0.0146$) despite balanced wins and losses ($8/8/4$). While \texttt{pmean} exhibits greater stability ($-0.022$ to $+0.019$), it trails \texttt{add} in both average score and win/loss ratio ($4/7/9$). As an unweighted linear sum, \texttt{add} provides a reliable baseline free of non-linear distortion. Because no alternative offers a statistically significant or systematic gain to justify added volatility, \texttt{add} is retained.

For \texttt{chan\_agg}, \texttt{max} is statistically indistinguishable from \texttt{softmax}, with all $20$ tuning series tying within $0.001$ VUS-PR. Although \texttt{mean} achieves a slightly higher overall average ($+0.0007$ VUS-PR; $6$ wins, $3$ losses), its per-series differences range from $-0.017$ to $+0.013$, with no statistically significant improvement ($p$-value $=0.154$). This variability reinforces the sensitivity concerns noted for $w_\ell$, $\beta$, and $\alpha$ in Section~\ref{sec:results-sensitivity}. Consequently, \texttt{softmax} is retained, as neither alternative demonstrates a significant or consistent advantage.

\subsection{Backbone Architecture Sensitivity}
\label{app:sensitivity-arch}

Five architectural constants of the masked-patch Transformer backbone were fixed prior to evaluation rather than jointly tuned: patch length $p$, depth $B$, width $D$, masking groups $G$, and mask block size $b$. To validate these choices, each constant was evaluated via single-variable grid sweeps on the $20$-series tuning set while holding all remaining hyperparameters at their deployed values ($w_\ell=1.0$, $\beta=1.0$, $\alpha=0.25$, $\gamma=1.5$, $\delta=1.5$). No candidate configuration outperformed the selected baseline ($p=16$, $B=4$, $D=128$, $G=4$, $b=1$), which was subsequently locked across all held-out evaluations. Table~\ref{tab:arch-constants} compares the top-performing alternative for each parameter against this baseline.

\begin{table}[htbp]
\centering
\small
\caption{Sensitivity of the five architectural constants fixed a priori (patch length $p$, depth $B$, width $D$, masking groups $G$, mask block $b$), evaluated one-at-a-time on the $20$-series tuning set while holding the remaining constants and all fusion coefficients at their deployed values. ``Best alternative vs.\ selected'' reports the single grid point with the highest mean VUS-PR among the non-selected values, its mean per-series $\Delta$ against the selected value, and a paired Wilcoxon signed-rank $p$-value; W/L/T counts series with a gain, loss, or tie at a $0.001$ VUS-PR threshold.  The selected configuration attains a mean VUS-PR of $0.6717$.
}
\label{tab:arch-constants}
\begin{tabular}{llcclc}
\toprule
Constant & Grid tested & Selected &
Best alternative ($\Delta$) & $p$-value & W/L/T \\
\midrule
$p$ (patch length)   & $\{8,16,32\}$    & $16$  & $32$: $-0.0347$ & $0.014$ & $4/14/2$ \\
$B$ (depth)          & $\{2,4,8\}$      & $4$   & $2$: $-0.0070$  & $0.891$ & $7/7/6$  \\
$D$ (width)          & $\{64,128,256\}$ & $128$ & $64$: $-0.0010$ & $0.984$ & $7/8/5$  \\
$G$ (masking groups) & $\{2,4,8\}$      & $4$   & $2$: $-0.0053$  & $0.225$ & $6/8/6$  \\
$b$ (mask block)     & $\{1,2,4\}$      & $1$   & $2$: $-0.0016$  & $0.395$ & $5/9/6$  \\
\bottomrule
\end{tabular}
\end{table}

Patch length is the sole architectural constant whose variation produces a statistically significant performance change. Both alternative patch resolutions induce notable drops ($p=8$: $\Delta=-0.0454$, $p$-value $=0.003$; $p=32$: $\Delta=-0.0347$, $p$-value $=0.014$), confirming that patch granularity critically influences the scale of extracted temporal features. Conversely, varying depth, width, masking groups, or mask block size yields no statistically significant performance change ($p$-values between $0.225$ and $0.984$), with relatively balanced win/loss ratios and small mean differences ($\leq 0.0070$ VUS-PR). Once an appropriate patch size is fixed, backbone performance remains robust across the tested architectural variations.

\end{document}